\documentclass[
	final,
	notitlepage,
	narroweqnarray,
	inline,
	twoside,
]{IEEEtran}

\usepackage{ieeefig}
\usepackage{amsmath}	
\usepackage{multirow}	
\usepackage{graphics}

\def\degree{^{\circ}}	

\begin{document}

\title{A GNSS Aided Initial Alignment Method for MEMS-IMU Based on Backtracking Algorithm and Backward Filtering}

\author{Xiaokang~Yang, Gongmin~Yan, Hao~Yang and Sihai~Li
}


\maketitle

\begin{abstract}
To obtain a high-accuracy position with SINS(Strapdown Inertial Navigation System), initial alignment needs to determine initial attitude rapidly and accurately. High-accuracy grade IMU(Inertial Measurement Uint) can obtain the initial attitude indenpendently, however, the low-accuracy grade gyroscope doesn't adapt to determine the heading angle, hence the initial attitude matrix will not be obtained. If using large misalignment angle model to estiamting heading angle, the convergence time will become much longer. For solving these two problems, a novel alignment algorithm combined backtracking algorithm and reverse navigation updating method with GNSS(Global Navigation Satellite System) aiding is proposed herein. The simulation and land vehicle test were finished to evaluate the alignment accuracy of the proposed algorithm. The horizontal misalignment is less than 2.3 arcmin and the heading misalignment is less than 10.1 arcmin in test. The proposed algorithm is a feasible and practical alignment method for low-cost IMU to obtain initial attitude in short term and large misalignment condition aided by GNSS.
\end{abstract}


\section{Introduction}
The strapdown initial navigation system (SINS) obtains precise attitude, velocity and position from angular rate and acceleration measured by gyroscopes and accelerometers. It is a dead-reckoning system, hence, the initial navigation information must be determined precisely at first. The alignment technology, that obtains initial attitude from gyroscopes and accelerometers output, is an indispensable part of navigation updating. Generally, a whole alignment process is composed with coarse aligning stage and fine aligning stage. An initial DCM (direction cosine matrix) will be calculated as the coarse aligning result in quasi-static base condition or swing base conditions. Britting \cite{Britting1971} proposed the analytic coarse alignment algorithm based on dual-vector, that is often used in stationary base condition. To finish aligning with latitude unknown, Yan \cite{Yan2020} calculates latitude with dual gravity vector and proposed the algorithm that obtain optical DCM with SVD (singular value decomposition). The SINS on vehicle and ship always aligns under the swing and vibrating conditions, hence Qin \cite{Qin2005} designd a clever aligning method by determining two inertial frame to avoid the disturbance. While Lian\cite{Lian2007} chose to filtering the disturbance in IMU outputs with multi-FIR-filters per-process. These algorithms are used to calculate a coarse attitude, and the fine initial attitude will be estimated after modeling the error model of SINS, estimating attitude errors of coarse aligning result, and analysing the state observablity degree. The error model is mainly divided into little-misalignment inertial model \cite{Fu2020} and large-misalignment nonlinear model \cite{Chang2019, Wen2020, Rahimi2020}. UKF (Unsented Kalman Filter) is usually used to estimate attitude error of nonlinear model, because large-misalignment error model of SINS is strong-nonlinearity. To make sure that the estimation results of attitude errors are all convergent in aligning period, the observablity of estimated states must be analysed according to the state error convariance, its eigvalues or eigvectors, and its singular value \cite{Yan2018, Ham1983, Sun2002, Huang2017}.

In the past decades, as the precision of MEMS-IMU(Micro-Electro-Mechanical System-IMU) continually increasing, all kinds of inertial navigation system based on MEMS-IMU have occurred in low-cost AHRS(Attitude Heading Reference System)\cite{Sheng2015}, small UAVs\cite{Jung2007}, the integrated navigation aided by GNSS\cite{Nourmohammadi2018}, indoor navigation\cite{Fan2017} and land vehicle navigation\cite{Georgy2011, Yang2014}. However, the traditional alignment algorithms researched for optic gyroscope INS(Initial Navigation System) are not valid anymore. The main problem of MEMS-IMU alignment is that the measurement precise of gyroscopes doesn't adapt to calculate initial heading. The static bias of MEMS gyroscope are much greater than the angular rate of earth rotation. Hence, neither the analytic alignment algorithm nor the inertial frame alignment algorithm cannot work out relatively accurate initial attitude. In this situation, longer aligning period is much necessary in the fine aligning stage. To solve the MEMS-IMU precisely aligning problem, many researchers propose different methods for all kinds of application situations. Yuan\cite{Yuan2015} proposed a indoor MEMS-IMU aligning method aided by magnetometers with UKF. For airborne MEMS INS, the transfer alignment method using main SINS or GNSS output as measurement information is wildly used\cite{Chu2017, Huang2020}. Xing\cite{Xing2018} used the rotation modulation technique to obtain precise alignment result on swing base, however the extra transposition mechanism need be installed on vehicle and the simple and small advantage will be no longer in existence. Wang\cite{Wang2017} realized in-flight alignment via UKF, but the same performance couldn't be obtained on a land vehicle in the same period because of the low observability of attitude error.

To solve these alignment problems of MEMS-IMU, a GNSS aided initial alignment method for MEMS-IMU is proposed herein. Two important methods, backtracking algorithm and backward filtering are combined in this novel alignment algorithm. Firstly, the nonlinear SINS error model with Euler platform error angle (EPEA), that represents the Euler angle between calculation frame and navigation frame, is used in attitude error estimation. It means the coarse aligning process is not necessary anymore and the attitude result can be estimated in any initial attitude value. Different from the data processing of traditional alignment, the whole data from gyroscopes and accelerometers need be stored and will be reprocessed many times. 
Using this MEMS-IMU alignment algorithm, the problems that coarse alignment result has large misalignment and the convergence time is too long for practical application will be effectively solved.
The proposed algorithm has several advantages as follows:
\par (1) The large misalignment SINS error model is used in the aligning process, hence the coarse alignment or an initial attitude matrix is not necessary. The precise attitude of vehicle can be estimated aided by GNSS with the proposed method in any initial attitude and any motion condition.
\par (2) Backtracking method is useful for estimating attitude in short-period alignment test. The limit IMU and GNSS data will be processed many times until the estimation result of attitude is convergent.
\par (3) The inverse navigation update and inverse Kalman filter update make the forward processing and backward processing be finished with the same program, and only need data be adjusted a bit.

\section{Initial Alignment based on Backtracking Filtering}

\subsection{Reverse navigation and its nonlinear error model}

For an initial alignment process, except the requirement of accuracy of error model, alignment time is another import standard to evaluate an alignment scheme. If reducing the time of alignment to realize fast alignment, the accuracy of alignment will be affected because of lack of necessary measurement. If prolonging the time of alignment to realize high-accuracy alignment, the requirement of fast alignment will not be met. To resolve this contradiction, we can firstly finish a fast alignment in a short time and store measurement date. Then, reuse the measurement data stored before to revise the alignment result with continuously finishing navigation calculation along the forward and reverse direction, until the accuracy doesn’t increase or reach the accuracy level that inertial navigation requires. For this target, the reverse navigation algorithm of SINS must be designed.

Assuming the sampling time of gyroscope and accelerator is $ T_s $. The forward navigation recursion algorithm used by navigation computer is shown in Eqs. (\ref{eq:fn_att}), (\ref{eq:fn_vn}) and (\ref{eq:fn_pos}).
\begin{equation}	\label{eq:fn_att}
\boldsymbol{C}_{b, k}^{n}=\boldsymbol{C}_{b, k-1}^{n}\left[\mathbf{I}+T_{s}\left(\boldsymbol{\omega}_{n b, k}^{b} \times\right)\right]
\end{equation}
\begin{equation}	\label{eq:fn_vn}
\begin{aligned}
\boldsymbol{v}_{k}^{n}=&\boldsymbol{v}_{k-1}^{n}+T_{s}\left[\boldsymbol{C}_{b, k-1}^{n} \boldsymbol{f}_{s f, k}^{b} \right. \\ & -\left.\left(2 \boldsymbol{\omega}_{i e, k-1}^{n}+\boldsymbol{\omega}_{e n, k-1}^{n}\right) \times \boldsymbol{v}_{k-1}^{n}+\boldsymbol{g}^{n}\right]
\end{aligned}
\end{equation}
\begin{subequations}	\label{eq:fn_pos}
	\begin{equation}
	L_{k}=L_{k-1}+\frac{T_{s} v_{N, k-1}^{n}}{R+h_{k-1}}
	\end{equation}
	\begin{equation}
	\lambda_{k}=\lambda_{k-1}+\frac{T_{s} v_{E, k-1}^{n} \sec L_{k-1}}{R+h_{k-1}}
	\end{equation}
	\begin{equation}
	h_{k}=h_{k-1}+T_{s} v_{U, k-1}^{n}
	\end{equation}
\end{subequations}
where $ \boldsymbol{C}_{b, k}^{n} $ is attitude matrix at $ t_k $, $ \boldsymbol{\omega}_{n b, k}^{b} $ is angular rate of body frame related to the navigation frame, $ \boldsymbol{v}_{k}^{n} $ is velocity at $ t_k $, $ \boldsymbol{f}_{s f, k}^{b} $ is specific force at $ t_k $, $ L_k $ is latitude, $ \lambda_{k} $ is longitude, $ h_k $ is altitude, and $ R $ is the earth radius. In this computing model, $ \boldsymbol{\omega}_{n b, k}^{b}=\boldsymbol{\omega}_{i b, k}^{b}-{\boldsymbol{C}_{b, k-1}^{n}}^\mathrm{T}\left(\boldsymbol{\omega}_{i e, k-1}^{n}-\boldsymbol{\omega}_{e n, k-1}^{n}\right) $, $ \boldsymbol{\omega}_{i e, k}^{n}=\left[\begin{array}{ccc}
0 & \boldsymbol{\omega}_{i e} \cos L_{k} & \boldsymbol{\omega}_{i e} \sin L_{k} \end{array} \right]^{\mathrm{T}} $ and $ \boldsymbol{\omega}_{e n, k}^{n}=\left[-\frac{v_{N, k}^{n}}{R+h_{k}} \frac{v_{E, k}^{n}}{R+h_{k}} \frac{v_{E, k}^{n} \tan L_{k}}{R+h_{k}} \right]^{\mathrm{T}} $, $ \left( k= 1,2,3,\cdots \right)  $ .

From the forward navigation algorithm, the reverse navigation algorithm, shown in Eqs. (\ref{eq:rn_att}), (\ref{eq:rn_vn}) and (\ref{eq:rn_pos}), can be deduced easily.
\begin{equation}	\label{eq:rn_att}
\begin{aligned}
\boldsymbol{C}_{b, k-1}^{n} & = \boldsymbol{C}_{b, k}^{n}\left[\boldsymbol{I}+T_{s}\left(\boldsymbol{\omega}_{n b, k}^{b} \times\right)\right]^{-1} \\ & \approx \boldsymbol{C}_{b, k}^{n}\left[\boldsymbol{I}-T_{s}\left(\boldsymbol{\omega}_{n b, k}^{b} \times\right)\right] \\ & \approx \boldsymbol{C}_{b, k}^{n}\left[\boldsymbol{I}+T_{s}\left(\tilde{\boldsymbol{\omega}}_{n b, k-1}^{b} \times\right)\right]
\end{aligned}
\end{equation}
\begin{equation}	\label{eq:rn_vn}
\begin{aligned}
\boldsymbol{v}_{k-1}^{n} =& \boldsymbol{v}_{k}^{n}-T_{s}\left[\boldsymbol{C}_{b, k-1}^{n} \boldsymbol{f}_{s f, k}^{b} \right. \\ 
& \left. -\left(2 \boldsymbol{\omega}_{i e, k-1}^{n}+\boldsymbol{\omega}_{e n, k-1}^{n}\right) \times \boldsymbol{v}_{k-1}^{n}+\boldsymbol{g}^{n}\right] \\
\approx & \boldsymbol{v}_{k}^{n}-T_{s}\left[\boldsymbol{C}_{b, k}^{n} \boldsymbol{f}_{s f, k-1}^{b}\right. \\ 
& \left. -\left(2 \boldsymbol{\omega}_{i e, k}^{n}+\boldsymbol{\omega}_{e n, k}^{n}\right) \times \boldsymbol{v}_{k}^{n}+\boldsymbol{g}^{n}\right]
\end{aligned}
\end{equation}
\begin{subequations}	\label{eq:rn_pos}
	\begin{equation}
	L_{k-1}=L_{k}-\frac{T_{s} v_{N, k-1}^{n}}{R+h_{k-1}} \approx L_{k}-\frac{T_{s} v_{N, k}^{n}}{R+h_{k}}
	\end{equation}
	\begin{equation}
	\lambda_{k-1}=\lambda_{k}-\frac{T_{s} v_{E, k-1}^{n} \sec L_{k-1}}{R+h_{k-1}} \approx \lambda_{k}-\frac{T_{s} v_{E, k}^{n} \sec L_{k}}{R+h_{k}}
	\end{equation}
	\begin{equation}
	h_{k-1}=h_{k}-T_{s} v_{U, k-1}^{n} \approx h_{k}-T_{s} v_{U, k}^{n}
	\end{equation}
\end{subequations}
where $ \tilde{\boldsymbol{\omega}}_{n b, k-1}^{b}=-\left[\boldsymbol{\omega}_{i b, k-1}^{b}-\boldsymbol{C}_{b, k}^{n}\left(\boldsymbol{\omega}_{i e, k}^{n}-\boldsymbol{\omega}_{e n, k}^{n}\right)\right] $.

Define $ \hat{\boldsymbol{ C }}_{b,m-j}^n = \boldsymbol{ C }_{b,j}^n $ ,$\hat{\boldsymbol{ v }}_{m-j}^n = -\boldsymbol{ v }_{j}^n $, $ \hat{L}_{m-j}^n = {L}_{j}^n $, $ \hat{\lambda}_{m-j}^n = {\lambda}_{j}^n $, $ \hat{h}_{m-j}^n = {h}_{j}^n $, $ \hat{\boldsymbol{ \omega }}_{ib,m-j}^b = -\boldsymbol{ \omega }_{ib,j}^b $, $ \hat{\boldsymbol{ f }}_{sf,m-j}^b = \boldsymbol{ f }_{sf,j}^b $, $ \hat{\boldsymbol{ \omega }}_{ie,m-j}^n = -\boldsymbol{ \omega }_{ie,j}^n $, $ \hat{\boldsymbol{ \omega }}_{en,m-j}^n = -\boldsymbol{ \omega }_{en-j}^n $, $ \hat{\boldsymbol{ \omega }}_{nb,m-j}^b = -\boldsymbol{ \omega }_{nb,j}^b $, and $ \hat{\boldsymbol{ \omega }}_{ie} = -\boldsymbol{ \omega }_{ie} $, $ \left( j=1,2,3,\cdots \right) $. Let $ p=m-k+1 $, the reverse navigation algorithm can be represented as
\begin{equation}	\label{eq:rn_att1}
\hat{\boldsymbol{C}}_{b, p}^{n}=\hat{\boldsymbol{C}}_{b, p-1}^{n}\left[\mathbf{I}+T_{s}\left(\hat{\boldsymbol{\omega}}_{n b, p}^{b} \times\right)\right]
\end{equation}
\begin{equation}	\label{eq:rn_vn1}
\begin{aligned}
\hat{\boldsymbol{v}}_{p}^{n}=&\hat{\boldsymbol{v}}_{p-1}^{n}-T_{s}\left[\hat{\boldsymbol{C}}_{b, p-1}^{n} \hat{\boldsymbol{f}}_{s f, p}^{b}\right. \\ 
& \left. -\left(2 \hat{\boldsymbol{\omega}}_{i e, p-1}^{n}+\hat{\boldsymbol{\omega}}_{e n, p-1}^{n}\right) \times \hat{\boldsymbol{v}}_{p-1}^{n}+\boldsymbol{g}^{n}\right]
\end{aligned}
\end{equation}
\begin{subequations}	\label{eq:rn_pos1}
	\begin{equation}
	\hat{L}_{p}=\hat{L}_{p-1}+\frac{T_{s} \hat{v}_{N, p-1}^{n}}{R+\hat{h}_{p-1}}
	\end{equation}
	\begin{equation}
	\hat{\lambda}_{p}=\hat{\lambda}_{p-1}+\frac{T_{s} \hat{v}_{E, p-1}^{n} \sec \hat{L}_{p-1}}{R+\hat{h}_{p-1}}
	\end{equation}
	\begin{equation}
	\hat{h}_{p}=\hat{h}_{p-1}+T_{s} \hat{v}_{U, p-1}^{n}
	\end{equation}
\end{subequations}
where
\begin{equation*}
\begin{aligned}
\hat{\boldsymbol{\omega}}_{i e, p}^{n} & = -\boldsymbol{\omega}_{i e, m-p}^{n} \\ 
&= -\left[\begin{array}{ccc}
0 & \omega_{i e} \cos L_{m-p} & \omega_{i e} \sin L_{m-p}
\end{array}\right]^{\mathrm{T}} \\ & = \left[\begin{array}{ccc}
0 & \hat{\omega}_{i e} \cos \hat{L}_{p} & \hat{\omega}_{i e} \sin \hat{L}_{p}
\end{array}\right]^{\mathrm{T}}
\end{aligned}
\end{equation*}
\begin{equation*}
\begin{aligned}
\hat{\boldsymbol{\omega}}_{e n, p}^{n} & = -\boldsymbol{\omega}_{e n, m-p}^{n} \\& = -\left[\begin{matrix}
	-{v_{N, m-p}^{n}}/\left(R+h_{m-p}\right) \\
	{v_{E, m-p}^{n}}/\left(R+h_{m-p}\right)  \\
	{v_{E, m-p}^{n} \tan L_{m-p}}/\left(R+h_{m-p}\right)
\end{matrix}\right] \\
&=\left[-\frac{\hat{v}_{N, p}^{n}}{R+\hat{h}_{p}} \frac{\hat{v}_{E, p}^{n}}{R+\hat{h}_{p}} \frac{\hat{v}_{E, p}^{n} \tan \hat{L}_{p}}{R+\hat{h}_{p}}\right]^{\mathrm{T}}
\end{aligned}
\end{equation*}
\begin{equation*}
\begin{aligned}
\hat{\boldsymbol{\omega}}_{n b, p}^{b}  =& \tilde{\boldsymbol{\omega}}_{n b, m-p}^{b} \\
 = & -\left[\hat{\boldsymbol{\omega}}_{i b, m-p}^{b}-\right. \\
   &\left.\boldsymbol{C}_{b, m-p+1}^{n}\left(\boldsymbol{\omega}_{i e, m-p+1}^{n}+\boldsymbol{\omega}_{e n, m-p+1}^{n}\right)\right] \\
 = & \hat{\boldsymbol{\omega}}_{i b, p}^{b}-\hat{\boldsymbol{C}}_{b, p-1}^{n}{ }^{\mathrm{T}}\left(\hat{\boldsymbol{\omega}}_{i e, p-1}^{n}+\hat{\boldsymbol{\omega}}_{e n, p-1}^{n}\right)
\end{aligned}
\end{equation*}

Comparing the reverse navigation algorithm in Eqs. (\ref{eq:rn_att1}), (\ref{eq:rn_vn1}) and (\ref{eq:rn_pos1}) with the forward navigation algorithm, the two algorithms have the same form. The sampling date can be used in reverse navigation algorithm, as long as inputting the negative values of gyroscope sampling and the angular rate of Earth rotation to the common navigation algorithm with the initial values set as $ \hat{\boldsymbol{C}}_{b_{0}}^{n}=\boldsymbol{C}_{b m}^{n} $, $ \hat{\boldsymbol{v}}_{0}^{n}=-\boldsymbol{v}_{m}^{n} $, $ \hat{L}_{0}=L_{m} $, $ \hat{\lambda}_{0}=\lambda_{m} $ and $ \hat{h}_{0}=h_{m} $.

To improve the accuracy of alignment, the parameter estimation methods like Kalman filter are used to estimate and compensate bias error of gyroscope and accelerator to improve accuracy indirectly, or to estimate and compensate attitude error for accurate alignment result directly. To estimate attitude error and bias error of gyroscope and accelerator during the period of reverse navigation, the corresponding error model need be built.

According to the analysis about reverse navigation algorithm, we know that the reverse algorithm has the same algorithm framework, so that the reverse update can be finished with forward algorithm after adjusting some input data and parameter. Hence, the error models of the two algorithms are much similar. 

Generally, attitude error is represented by the EPEA to construct nonlinear error model of SINS. Let $ \boldsymbol{\alpha} = [\alpha_x \quad \alpha_y \quad \alpha_z]^{\mathrm{T}} $ denotes EPEA. The attitude error represents the rotation from ideal navigation frame $ n $ to real navigation frame $ n' $. According to the calculation rule of DCM, the transformation matrix from frame $ n $ to frame $ n' $ is
\begin{equation}
\begin{aligned}
\boldsymbol{C}_n^{n'} =&  \left[ \begin{array}{c}
\mathrm{c} \alpha_{y} \mathrm{c} \alpha_{z}-\mathrm{s} \alpha_{y} \mathrm{s} \alpha_{x} \mathrm{s} \alpha_{z} \\
-\mathrm{c} \alpha_{x} \mathrm{s} \alpha_{z} \\
\mathrm{s} \alpha_{y} \mathrm{c} \alpha_{z}+\mathrm{c} \alpha_{y} \mathrm{s} \alpha_{x} \mathrm{s} \alpha_{z}
\end{array} \right. \\
& \left. \begin{array}{rl}
\mathrm{c} \alpha_{y} \mathrm{s} \alpha_{z}+\mathrm{s} \alpha_{y} \mathrm{s} \alpha_{x} \mathrm{c} \alpha_{z} & -\mathrm{s} \alpha_{y} \mathrm{c} \alpha_{x} \\
\mathrm{c} \alpha_{x} \mathrm{c} \alpha_{z} & \mathrm{s} \alpha_{x} \\
\mathrm{s} \alpha_{y} \mathrm{s} \alpha_{z}-\mathrm{c} \alpha_{y} \mathrm{s} \alpha_{x} \mathrm{c} \alpha_{z} & \mathrm{c} \alpha_{y} \mathrm{c} \alpha_{x}
\end{array} \right]
\end{aligned}
\end{equation}
where $ \mathrm{s}i $ denotes $ \sin(i) $, $ \mathrm{c}i $ denotes $ \cos(i) $. And the corresponding nonlinear error model is shown as
\begin{subequations}	\label{eq:errorModel}
\begin{equation}	\label{eq:err_alpha}
\dot{\boldsymbol{\alpha}}=\boldsymbol{C}_{\omega}^{-1}\left[\left(\mathbf{I}-\boldsymbol{C}_{n}^{n^{\prime}}\right) \tilde{\boldsymbol{\omega}}_{i n}^{n}+\boldsymbol{C}_{n}^{n^{\prime}} \delta \boldsymbol{\omega}_{i n}^{n}-\boldsymbol{C}_{b}^{n^{\prime}} \delta \boldsymbol{\omega}_{i b}^{b}\right]
\end{equation}
\begin{equation}	\label{eq:err_vn}
\begin{aligned}
\delta \dot{\boldsymbol{v}}^{n}
=&\left[\mathbf{I}-\left(\boldsymbol{C}_{n}^{n^{\prime}}\right)^{\mathrm{T}}\right] \boldsymbol{C}_{b}^{n^{\prime}} \tilde{\boldsymbol{f}}_{s f}^{b}+\left(\boldsymbol{C}_{n}^{n^{\prime}}\right)^{\mathrm{T}} \boldsymbol{C}_{b}^{n^{\prime}} \delta \boldsymbol{f}_{s f}^{b} \\
&-\left(2 \delta \boldsymbol{\omega}_{i e}^{n}+\delta \boldsymbol{\omega}_{e n}^{n}\right) \times \tilde{\boldsymbol{v}}^{n}-\left(2 \boldsymbol{\omega}_{i e}^{n}+\boldsymbol{\omega}_{e n}^{n}\right) \times \delta \boldsymbol{v}^{n} \\ 
&+\left(2 \delta \boldsymbol{\omega}_{i e}^{n}+\delta \boldsymbol{\omega}_{e n}^{n}\right) \times \delta \boldsymbol{v}^{n}+\delta \boldsymbol{g}^{n}
\end{aligned}
\end{equation}
\begin{equation}	\label{eq:dlat}
{\delta \dot{L}=\frac{1}{R+h} \delta v_{N}-\frac{v_{N}}{\left(R+h\right)^{2}} \delta h}
\end{equation}
\begin{equation}	\label{eq:dlng}
{\delta \dot{\lambda}=\frac{\sec L}{R+h} \delta v_{E}+\frac{v_{E} \sec L \tan L}{R+h} \delta L-\frac{v_{E} \sec L}{\left(R+h\right)^{2}} \delta h}
\end{equation}
\begin{equation}	\label{eq:dhgt}
{\delta \dot{\boldsymbol{h}} = \delta \boldsymbol{v}_U}
\end{equation}
\end{subequations}
where $\boldsymbol{C}_{\omega}$ is the transfer matrix from angular rate of computing frame to the derivation of EPEA. Its detail form is
\begin{equation}
	\boldsymbol{C}_{\omega} = \left[\begin{matrix}
		\cos(\alpha_{y}) & 0 & -\sin(\alpha_{y})\cos(\alpha_{x}) \\
				0          & 1 & \sin(\alpha_{x})  \\
		\sin(\alpha_{y}) & 0 & \cos(\alpha_{y})cos(\alpha_{x})
	\end{matrix}\right]
\end{equation}

Let $ \boldsymbol{\omega}_{ib}^b $ denote real angular rate of  vehicle, $ \delta \boldsymbol{\omega}_{ib}^b $ is bias error vector of triaxial gyroscope, then the input angular rate of reverse navigation algorithm is given by
\begin{equation} \label{eq:rwbib}
\tilde{\boldsymbol{\omega}}_{ib}^b = -\boldsymbol{\omega}_{ib}^b - \delta \boldsymbol{\omega}_{ib}^b
\end{equation}
where $ -(\boldsymbol{\omega}_{ib}^b) $ is regarded as real input for reverse algorithm. In Eq. (\ref{eq:rwbib}), the bias error $ \delta \boldsymbol{\omega}_{ib}^b $ is subtracted from real value, instead of added on real value of angular rate. So, the transition process of gyroscope bias is different. The attitude error model of reverse navigation algorithm is given by
\begin{equation}	\label{eq:rn_alpha1}
\dot{\boldsymbol{\alpha}}=\boldsymbol{C}_{\omega}^{-1}\left[\left(\mathbf{I}-\boldsymbol{C}_{n}^{n^{\prime}}\right) \tilde{\boldsymbol{\omega}}_{i n}^{n}+\boldsymbol{C}_{n}^{n^{\prime}} \delta \boldsymbol{\omega}_{i n}^{n}+\boldsymbol{C}_{b}^{n^{\prime}} \delta \boldsymbol{\omega}_{i b}^{b}\right]
\end{equation}

To make the error models of reverse navigation and normal navigation have same construct, the negative value of gyroscope measurement error   is substituted in Eq. (\ref{eq:rn_alpha1}). Its differential equation is given by
\begin{equation}	\label{eq:rn_alpha2}
\dot{\boldsymbol{\alpha}}=\boldsymbol{C}_{\omega}^{-1}\left[\left(\mathbf{I}-\boldsymbol{C}_{n}^{n^{\prime}}\right) \tilde{\boldsymbol{\omega}}_{i n}^{n}+\boldsymbol{C}_{n}^{n^\prime} \delta \boldsymbol{\omega}_{i n}^{n} -\boldsymbol{C}_{b}^{n^\prime}\left(-\delta \boldsymbol{\omega}_{i b}^{b}\right)\right]
\end{equation}

When establishing a Kalman filter according to Eq. (\ref{eq:rn_alpha2}) to estimate attitude error and gyroscope error, the estimated result of gyroscope error is the negative value of real gyroscope error. The normal navigation error model can be used to establish Kalman filter for reverse navigation algorithm, as long as the gyroscope error is handled correctly when setting initial value and variance, and getting estimated result. The velocity and position error model of reverse navigation algorithm is the same as the nonlinear error model shown in Eqs. (\ref{eq:err_vn})$ \sim $(\ref{eq:dhgt}).

The reverse navigation algorithm and the considered error model have been established. Hence, we can update navigation result and aligning Kalman filter by normal navigation algorithm and normal error model from the end point to the initial point. IMU data can be calculated as forward and backward, so a group of data can be reused many times until the attitude information is fully mined out from measurement data.

\subsection{Backtracking filtering with UKF}

In conventional fine alignment algorithm, both norm Kalman filter and nonlinear Kalman filter finish data process as forward direction. Although the backtracking algorithm has been applied to solve problems about alignment, the direction of filtering is still forward. To realize reusing recorded data to estimate misalignment angles and SIMU errors with Kalman filter, the backtracking filter method is proposed herein. Firstly, an UKF filter is deigned according to nonlinear error model of forward navigation algorithm. Then the backward filter will be designed to constitute the whole backtracking filtering algorithm with UKF.

From Eq. (\ref{eq:errorModel}), the state vector is set as
\begin{equation} 	\label{eq:stateVector}
\boldsymbol{X} = [\begin{matrix}
\boldsymbol{ \alpha }^\mathrm{T} & (\delta\boldsymbol{v}^n)^\mathrm{T} & \delta \boldsymbol{p} & ( \boldsymbol{\varepsilon}^b)^\mathrm{T} & ( \boldsymbol{\nabla}^b)^\mathrm{T}
\end{matrix}]^\mathrm{T}
\end{equation}
where $ \delta \boldsymbol{p} = [\begin{matrix}
\delta L & \delta \lambda & \delta h
\end{matrix}] $ is position error vector consist of latitude error $ \delta L $, longitude error $ \delta \lambda $ and altitude error $ \delta h $. $ \boldsymbol{\varepsilon}^b $ and $ \delta \boldsymbol{\nabla}^b $ are bias error of gyroscope and accelerometer respectively.

According to the error equations in Eq. (\ref{eq:errorModel}) and $ \delta \boldsymbol{\omega}^b_{ib} = \boldsymbol{\varepsilon}^b + \boldsymbol{w}^b $, the differential of state vector is given by
\begin{equation}
\dot{\boldsymbol{X}} = \boldsymbol{f}(\boldsymbol{X}) + \boldsymbol{g}(\boldsymbol{X}) \boldsymbol{w}^b
\end{equation}

With velocity in navigation frame and position from GPS, the measurement equation is a linear equation, which is given by 
\begin{equation}
\boldsymbol{X} = \boldsymbol{H}\boldsymbol{X} + \boldsymbol{v}
\end{equation}
where
\begin{equation}
\boldsymbol{H} = \left[\begin{matrix}
\boldsymbol{\mathrm{0}}_{3 \times 3} & \boldsymbol{\mathrm{I}}_{3 \times 3} & \boldsymbol{\mathrm{0}}_{3 \times 3} & \boldsymbol{\mathrm{0}}_{3 \times 6} \\ \boldsymbol{\mathrm{0}}_{3 \times 3} & \boldsymbol{\mathrm{0}}_{3 \times 3} & \boldsymbol{\mathrm{I}}_{3 \times 3} & \boldsymbol{\mathrm{0}}_{3 \times 6}
\end{matrix} \right]
\end{equation}
and $ \boldsymbol{ v } $ is measurement noise vector.

It is clear that the time-update equation of state space model is nonlinear and measurement-update equation is linear. Hence, nonlinear Kalman filter will be applied in alignment. The time-update process need to use UT (Unscented Transformation) method to finish updating of state vector and its variance. And the measurement-update process is similar with that in standard Kalman filter. The state vector in Eq. (\ref{eq:stateVector}) can be estimated based on nonlinear error model of SINS with UKF.

Let the start time and stop time of recorded IMU data denote as $ t_0 $ and $ t_1 $. After a forward UKF, we will have the last estimation result of $ \boldsymbol{X} $
\begin{equation} 	\label{eq:stateVector_t2}
\boldsymbol{X}_{t_1} = [\begin{matrix}
\boldsymbol{ \alpha }_{t_1}^\mathrm{T} & (\delta\boldsymbol{v}^n)_{t_1}^\mathrm{T} & \delta \boldsymbol{p}_{t_1} & ( \boldsymbol{\varepsilon}^b)_{t_1}^\mathrm{T} & ( \boldsymbol{\nabla}^b)_{t_1}^\mathrm{T}
\end{matrix}]^\mathrm{T}
\end{equation}


Compare the linear and nonlinear error model of SINS, the sensor error of gyroscope $ \delta \boldsymbol{\omega}^b_{ib} $ is the only differentia. Let $ \boldsymbol{X}^\prime_{t_1} $ denote the initial state vector of backward filter. Before the backward filtering,  $ \boldsymbol{X}^\prime_{t_1} $ should be set as
\begin{equation} 	\label{eq:stateVector_t2_}
\boldsymbol{X}^\prime_{t_1} = [\begin{matrix}
\boldsymbol{ \alpha }_{t_1}^\mathrm{T} & (\delta\boldsymbol{v}^n)_{t_1}^\mathrm{T} & \delta \boldsymbol{p}_{t_1} & -(\boldsymbol{\varepsilon}^b)_{t_1}^\mathrm{T} & ( \boldsymbol{\nabla}^b)_{t_1}^\mathrm{T}
\end{matrix}]^\mathrm{T}
\end{equation}
and the rest of UKF will not be change. In the same way, the negative vector of estimation result of $ \boldsymbol{\varepsilon}^b $ will be token to reconstructed state vector for next UKF that updates as reverse direction.


\subsection{A novel scheme of GPS-assistant inertial alignment}

\begin{figure}[!ht]
	\centering
	\includegraphics[width=7cm]{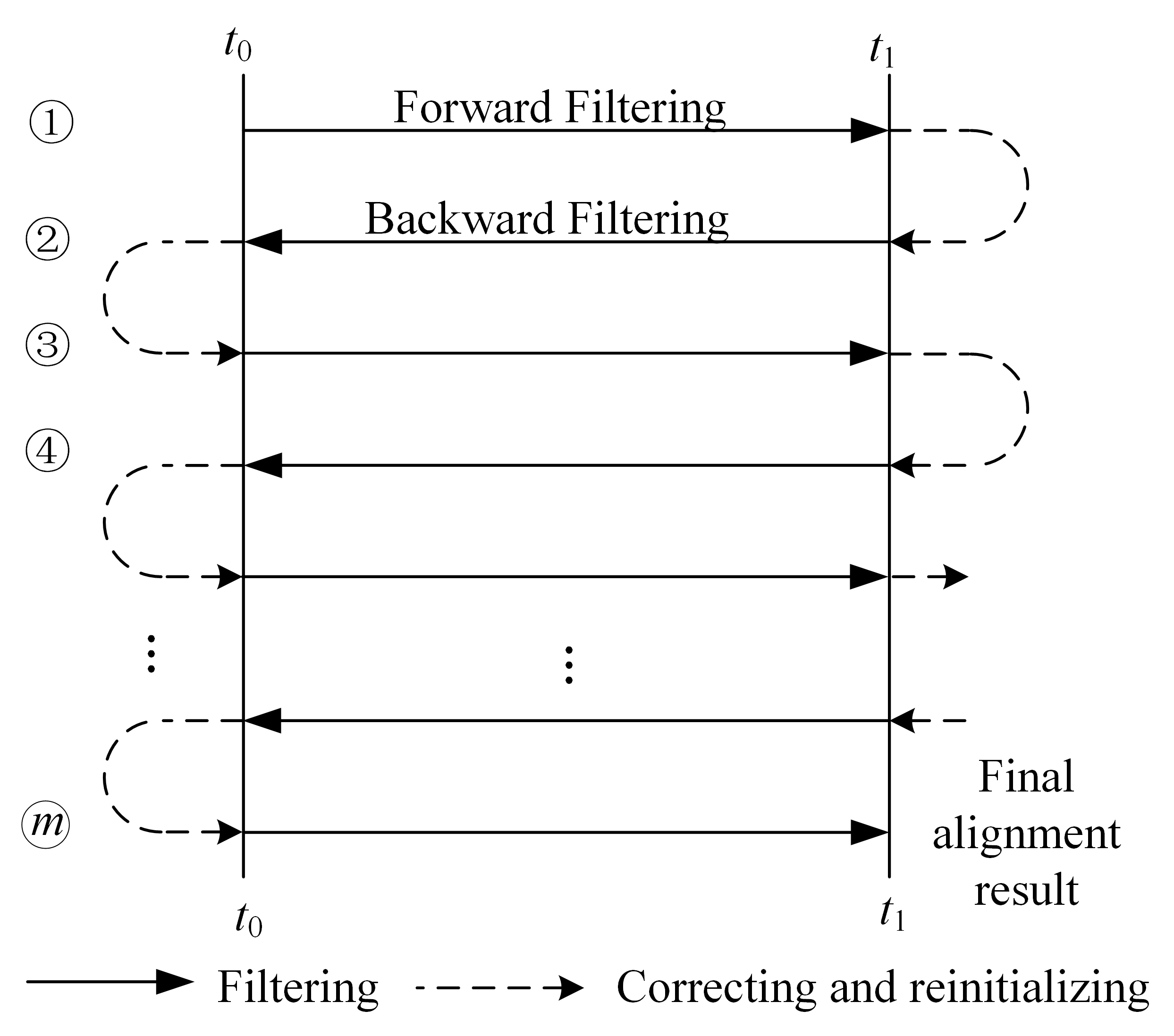}
	\caption{Diagram of alignment scheme of GPS-assistant alignment method based on backtracking algorithm} 
	\label{fig:dig}
\end{figure}

The simple diagram of alignment scheme of proposed method is shown in Figure \ref{fig:dig}. In the diagram, a whole alignment scheme is divided into many forward and reverse data processes, that uses nonlinear Kalman filter to estimate the misalignment angle and bias error of SIMU. These data processes are denoted as \textcircled{1}, \textcircled{2}, ... and \textcircled{\textit{m}}. If the accuracy of IMU is enough high to work out the velocity, the alignment in the inertial frame  introduced in \cite{Xing2018} can be used to finish determining an approximate initial attitude. 
To use backtrack algorithm in alignment, the whole IMU aligning data is proposed and stored.  The initial time point is denoted as $t_0$ and the finished time point is denoted as $t_1$. The backtracking alignment, which is composed with backward navigation and backward filtering, can improve the alignment accuracy and convergence rate by increasing the aligning stage.

\begin{figure}[!ht]
	\centering
	\includegraphics[width=7cm]{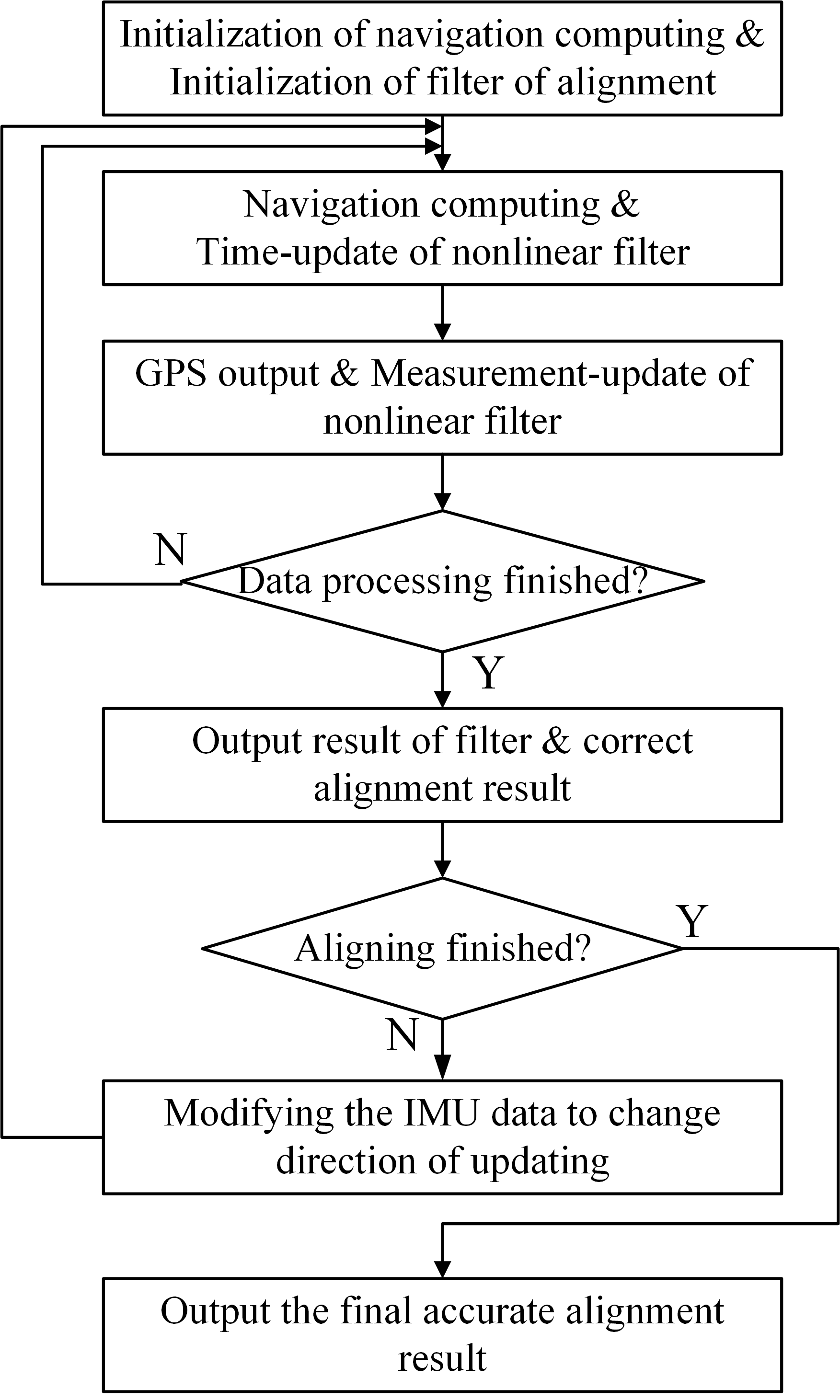}
	\caption{The flow diagram of whole fine alignment stage}
	\label{fig:dig_fine}
\end{figure}

In each stage, every filtering process is established based on the nonlinear error model of SINS to improve the accuracy of error model. After a normal forward navigation computing and fine alignment based on EKF or UKF, the next processes of navigation computing and fine alignment are finished from $ t_1 $ to $ t_0 $, according the reverse navigation algorithm and its nonlinear error model introduced in this section. After finish the forward and reverse process many times, the estimating result of misalignment angle will converge to an accurate result closed to real misalignment angle. The processing procedure of the whole alignment is represented as the flow diagram shown in Figure \ref{fig:dig_fine}.

The proposed aligning process is much easy to realize. Firstly, initialize all kinds of parameters of nonlinear filter. Next, finish navigation updating, time-update of nonlinear filter and measurement-update of nonlinear filter in sequence. If all sensors data, that stored during coarse alignment, is processed from $ t_0 $ to $ t_1 $, the first estimation result of misalignment angle will be obtained from process \textcircled{1}. And in the next process, navigation computing and alignment nonlinear filtering will be updated from $ t_1 $ to $ t_0 $. The result of forward processing need be transformed to the initial parameters of reverse navigation updating and  reverse nonlinear filter. With repeating the forward updating and reverse updating many times, the state vector representing misalignment angle will converge to real value endlessly.

In short, this novel scheme of alignment integrate the accurate nonlinear model of SINS, nonlinear Kalman filter, reverse navigation algorithm and backtracking alignment algorithm. All advantages of these excellent algorithm are used to improve accuracy of alignment and reduce alignment time. It is feasible and reasonable to exchange the accuracy of alignment with computing power in the current technological background, although the proposed alignment algorithm and scheme need finish more navigation updating and nonlinear Kalman filtering.

\section{Evaluation of Algorithm Performance}

\subsection{Simulation of alignment}

To verify the proposed alignment method and compare its alignment result with other alignment methods, a 600s trajectory of vehicle is simulated to align on moving base. The alignment simulations with different alignment algorithms were finished in the same initial misalignment condition. The specifications of SIMU is shown in Table~\ref{tab:sim}.

\begin{table}[!ht]
	\centering
	\caption{Specifications of SIMU} \label{tab:sim}
	\begin{tabular}{lll}
		\hline
		Sensor & Error parameter & Value \\
		\hline
		\multirow{2}{*}{Gyroscope} & Bias & $ 1\degree/\mathrm{h} $ \\
		 & Random walk & $ 0.1 \degree/\sqrt{\mathrm{h}} $ \\
		\multirow{2}{*}{Accelerometer} & Bias & $ 2\mathrm{mg} $ \\
		 & Random walk & $ 1\mathrm{mg}/\sqrt{\mathrm{Hz}} $\\
		\hline
	\end{tabular}
\end{table}

Because we need to compare linear error model and nonlinear error model and to compare backtracking algorithm and non-backtracking algorithm, 4 different alignment methods listed in Table \ref{tab:sim_4alist} are used to align with the same trajectory data.
\begin{table}[!ht]
	\centering
	\caption{Four algorithms used in simulation} \label{tab:sim_4alist}
		\begin{tabular}{llp{5cm}}
		\hline
		Num. & Abbr. & Alignment method \\
		\hline
		1 &	LM	 & 	Linear model without backtracking algorithm \\
		2 &	LMBT &	Linear model with backtracking algorithm \\
		3 &	NM	 &	Nonlinear model without backtracking algorithm \\
		4 &	NMBT &	Nonlinear model with backtracking algorithm \\
		\hline
		\end{tabular}
\end{table}

\begin{figure}[!ht]
	\centering
	\includegraphics[width=\columnwidth]{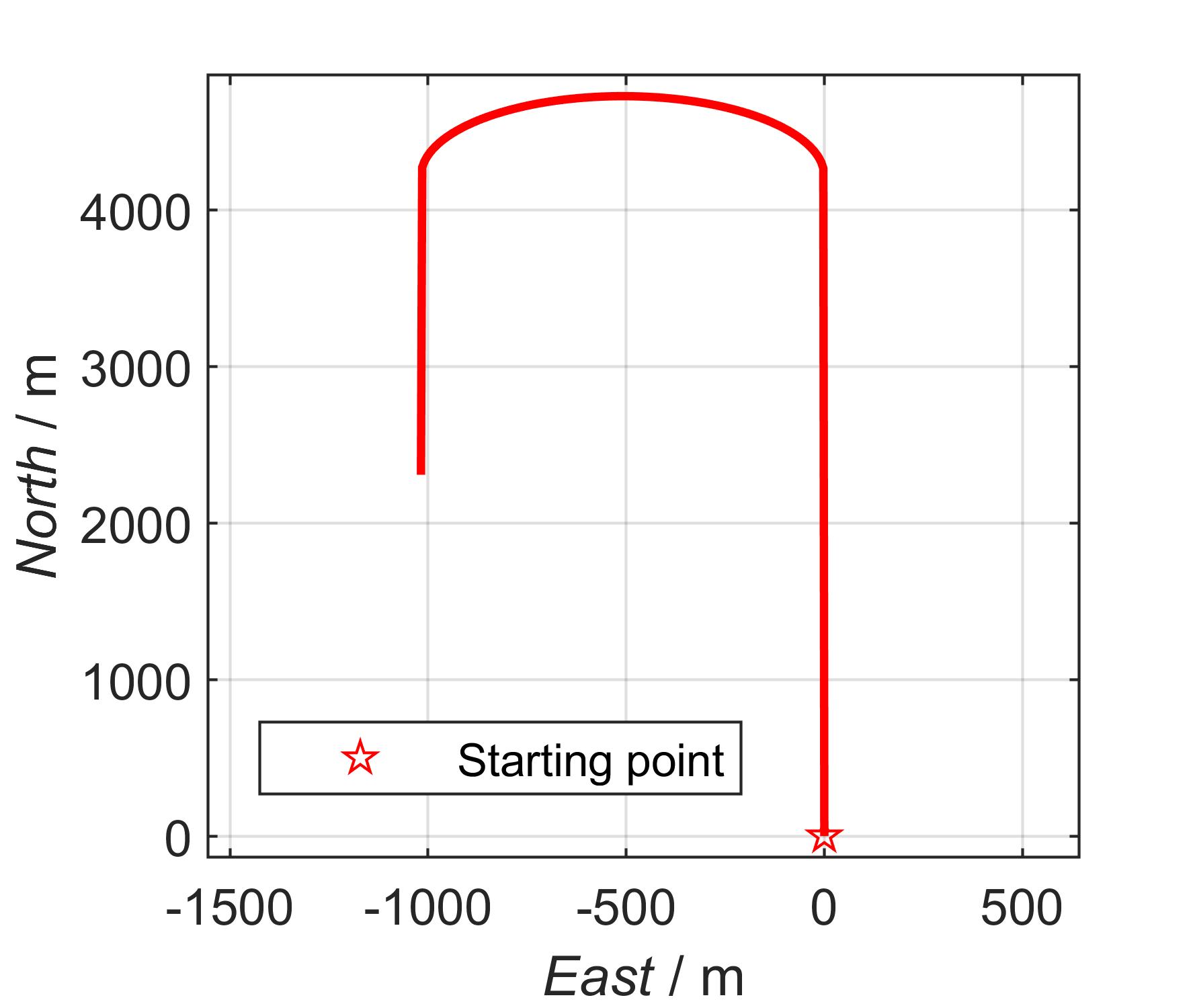}
	\caption{Trajectory of vehicle in simulation}
	\label{fig:trj1}
\end{figure}

\begin{figure}[!ht]
	\centering
	\includegraphics[width=\columnwidth]{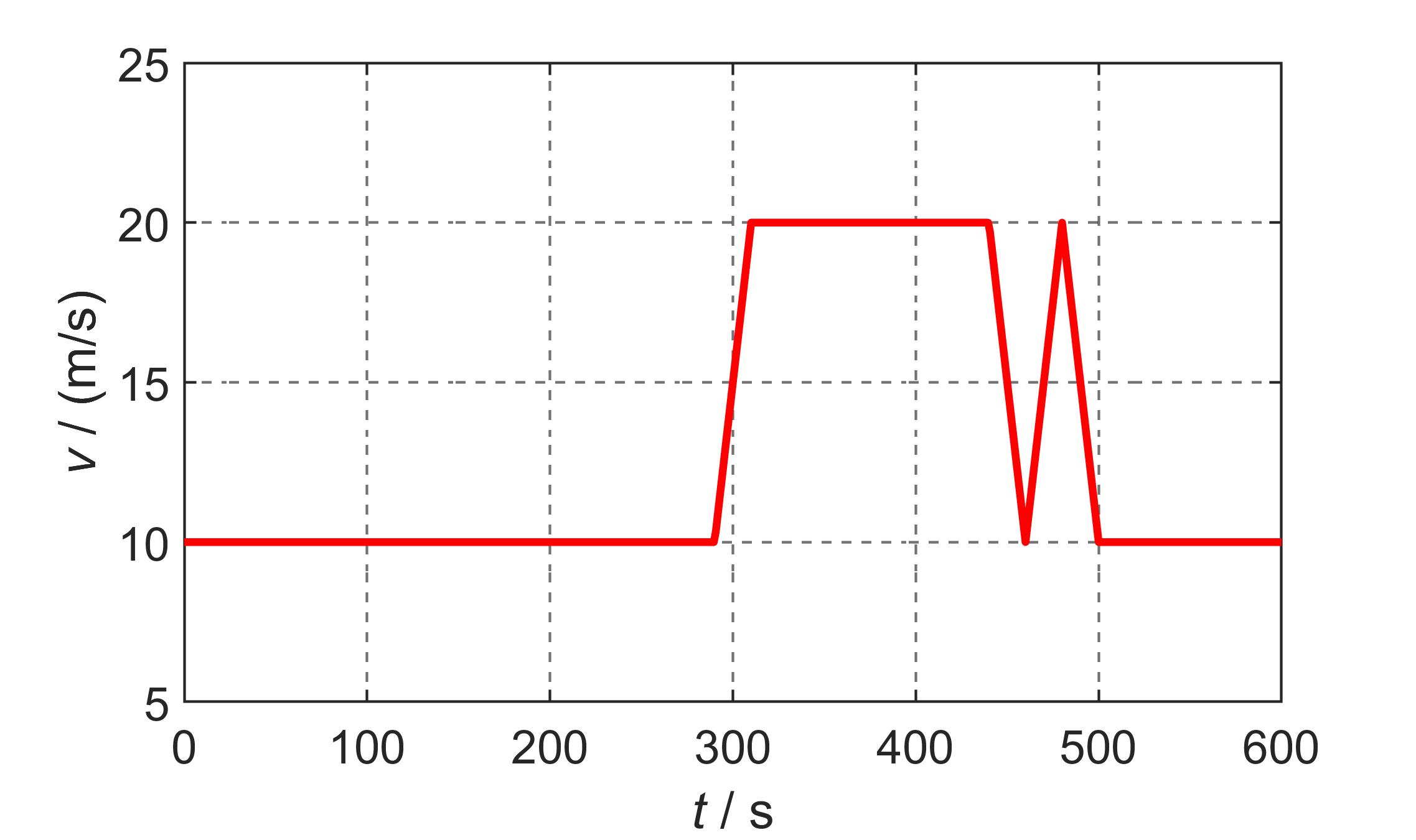}
	\caption{Curve of velocity in simulation}
	\label{fig:simvn}
\end{figure}

The simulation trajectory is shown as Figure \ref{fig:trj1}. And the velocity variance is shown in Figure \ref{fig:simvn}. The initial velocity is $ 10 \mathrm{m/s} $. After 290s uniform line motion, the vehicle accelerates to $20 \mathrm{m/s}$. A $ 180 \degree $ rotation is finished after accelerating. To improve the observability degree of misalignment angels, the vehicle finfish three acceleration or deceleration maneuvers before 500s.

All alignment method shown in Table \ref{tab:sim_4alist} are used to finish alignment and obtain estimation value of misalignment angles. If alignment algorithm is based on backtracking algorithm, the stored SIMU data will be processed 3 times, and the estimation results in the last filtering will compare with other alignment algorithms. 

\begin{figure}[!ht]
	\centering
	\includegraphics[width=\columnwidth]{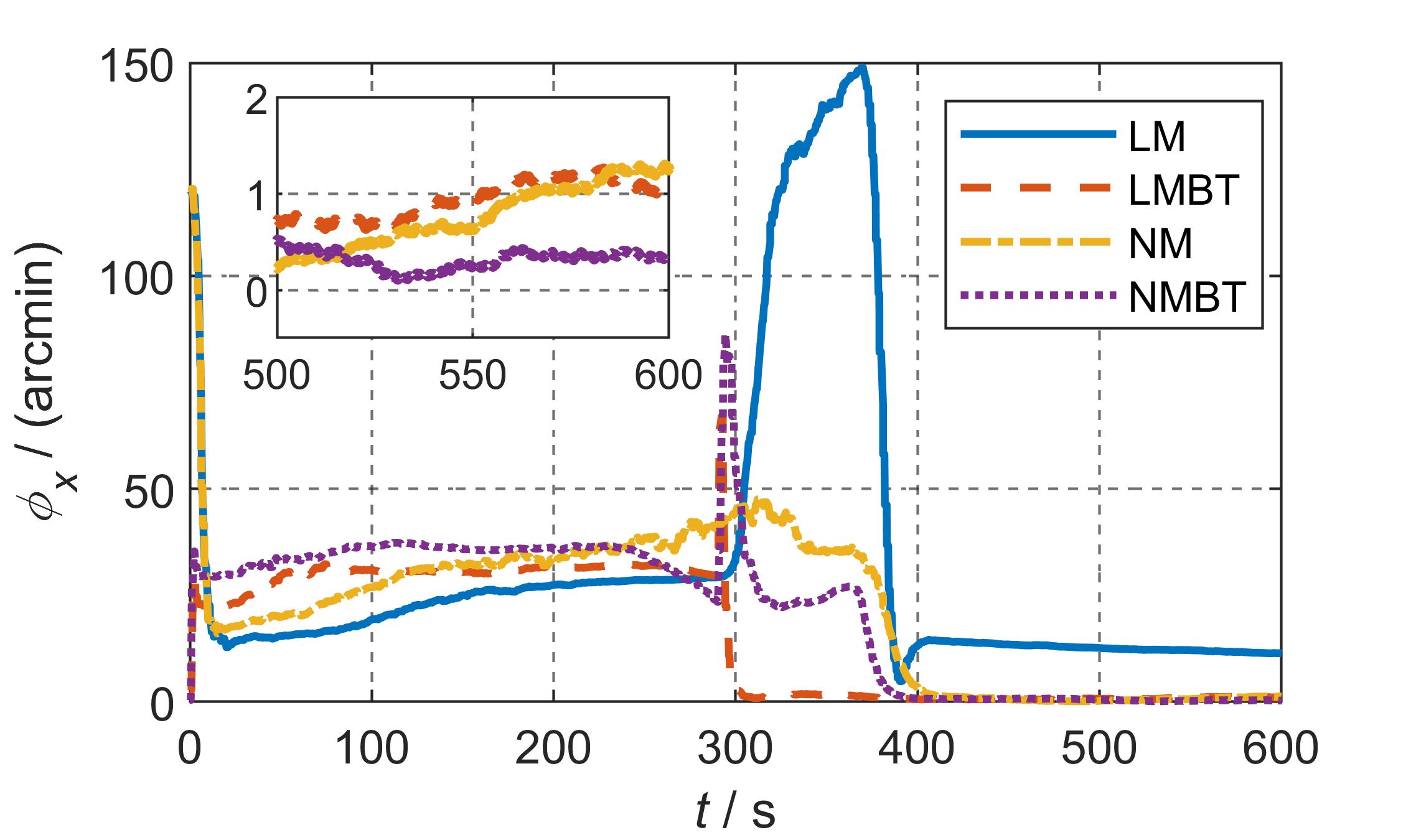}
	\caption{Estimation errors of \textit{x}-axis misalignment angle in simulation}
	\label{fig:sim_x}
\end{figure}

\begin{figure}[!ht]
	\centering
	\includegraphics[width=\columnwidth]{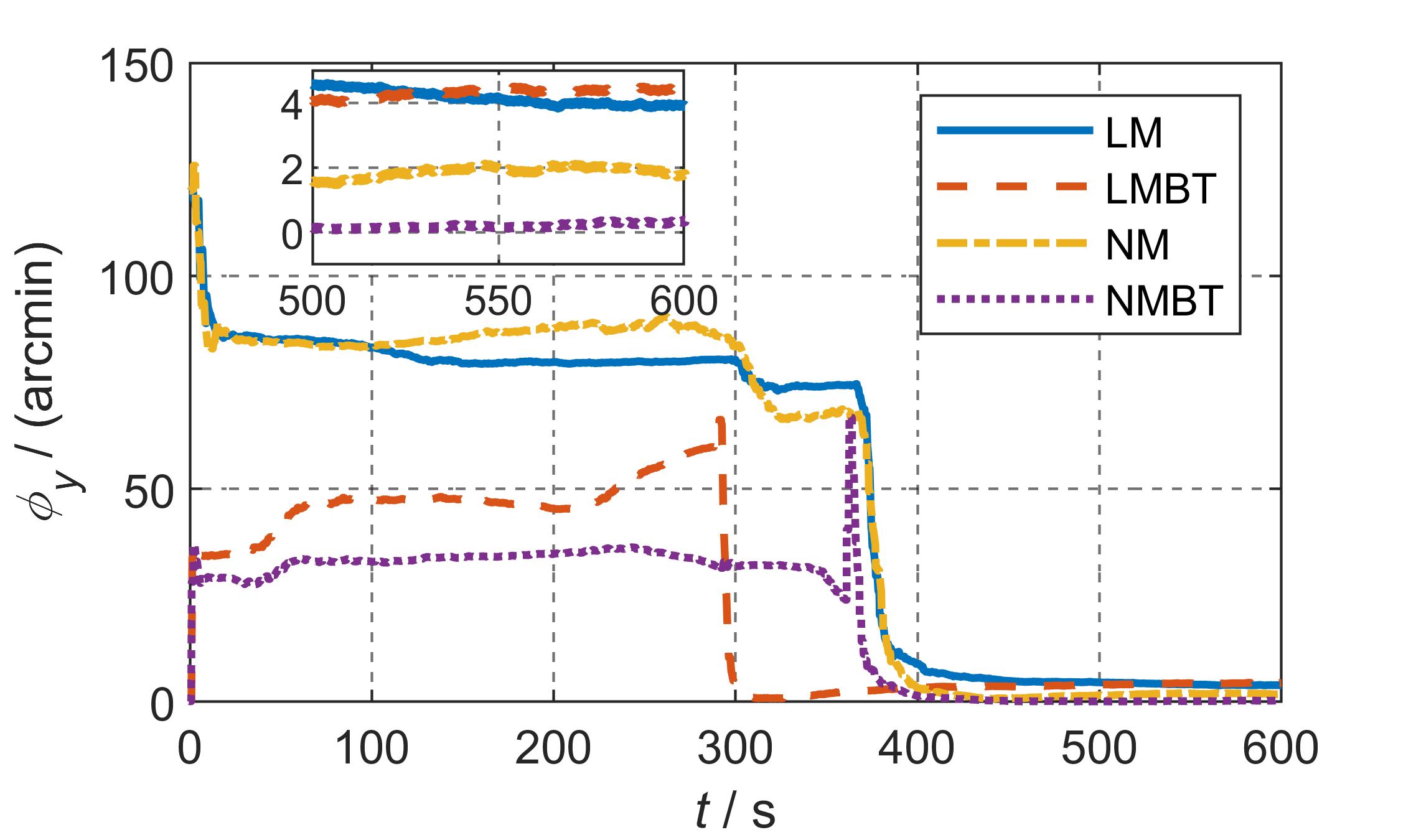}
	\caption{Estimation errors of \textit{y}-axis misalignment angle in simulation}
	\label{fig:sim_y}
\end{figure}

\begin{figure}[!ht]
	\centering
	\includegraphics[width=\columnwidth]{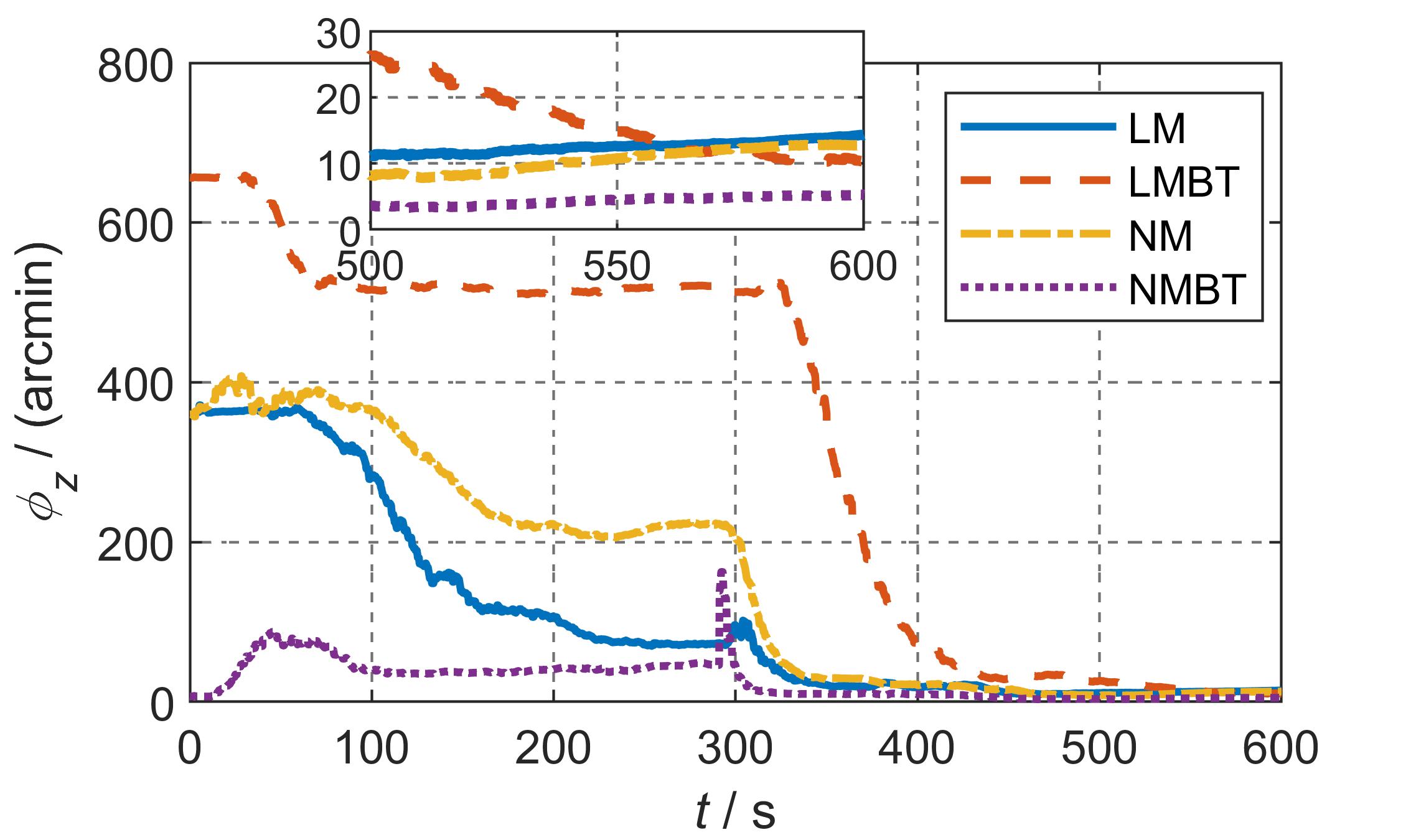}
	\caption{Estimation errors of \textit{z}-axis misalignment angle in simulation}
	\label{fig:sim_z}
\end{figure}

Firstly, the simulations in the little misalignment angle condition were finished as $ \boldsymbol{\alpha} = \left[ \begin{matrix}
	1\degree & 1\degree & 3\degree 
\end{matrix} \right] $. To compare accuracy of algorithms, the estimation results of misalignment angles are obtained with carrying out 30 Monte-Carlo simulation tests. The RMS estimation errors of misalignment angles of these alignment algorithms are shown in Figures \ref{fig:sim_x}, \ref{fig:sim_y} and \ref{fig:sim_z}. It is clear that the estimation performance of estimation results of 3 misalignment angles are not good in the first 290s, because the vehicle is in uniform linear motion, which offers limit contribution for estimating states of Kalman filter. After a series of maneuver consist of $ 180\degree $ rotation, acceleration and deceleration, the estimation results of misalignment angle converge to near the real misalignment angles. The partial enlarged figures, from 500s to 600s, of Figures \ref{fig:sim_x}, \ref{fig:sim_y} and \ref{fig:sim_z} are located in every original figure. It indicates that the alignment method based on nonlinear error model with backtracking algorithm (NMBT) has the highest accuracy in estimation of 3 misalignment angle. In the result of x-axis misalignment angle, the estimation error of NMBT algorithm is smallest. And the estimation error of NM and LMBT algorithms are greater than NMBT algorithm, but far smaller than the LM algorithm. In the result of y-axis misalignment angle, the estimation accuracy from high to low is NMBT algorithm, NM algorithm, LM algorithm and LMBT algorithm. The z-axis misalignment angle, representing the error of yaw, is one of the important evaluation parameters of SINS. In these alignment algorithms, the NMBT algorithm has highest accuracy. The other algorithms are similar and greater than NMBT algorithm. 

The estimation errors of these alignment algorithms at the end of simulation are listed in Table \ref{tab:simerr}. It is concluded that the NMBT algorithm has the highest accuracy in the simulation condition. The advantages of accurate error model and repeatedly using SIMU data improve estimation accuracy of alignment significantly. Modify the conventional alignment algorithm based on either nonlinear error model or backtracking algorithm can improve alignment accuracy, but there still exists a large accuracy gap between NMBT.

\begin{table}[!ht]
	\centering
	\caption{Estimation error (RMS) of misalignment angle of different alignment algorithms} \label{tab:simerr}
	\begin{tabular}{lrrr}
		\hline
		Alignment algorithm & $ \phi_x (^\prime) $ & $ \phi_y (^\prime) $ & $ \phi_z (^\prime) $ \\
		\hline
		LM   & 11.3760 & 3.9011 & 14.3020  \\
		LMBT & 1.0420  & 4.4186 & 10.2190  \\
		NM   & 1.2418  & 1.7750 & 12.8032  \\
		NMBT & 0.3123  & 0.3429 & 5.2764   \\
		\hline
	\end{tabular}
\end{table}

\begin{figure}[!ht]
	\centering
	\includegraphics[width=\columnwidth]{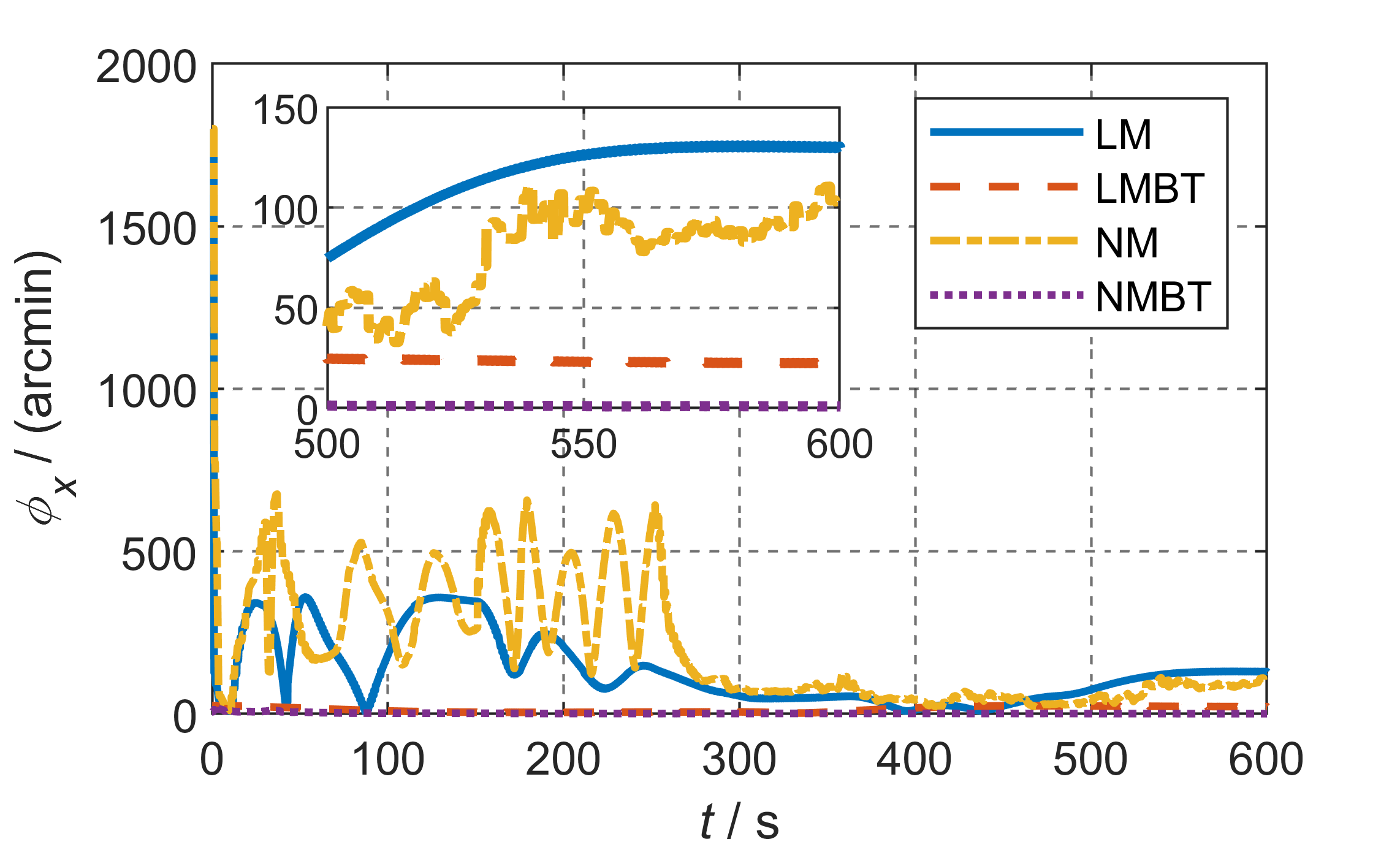}
	\caption{Estimation errors of \textit{x}-axis misalignment angle in simulation}
	\label{fig:sim2_x}
\end{figure}

\begin{figure}[!ht]
	\centering
	\includegraphics[width=\columnwidth]{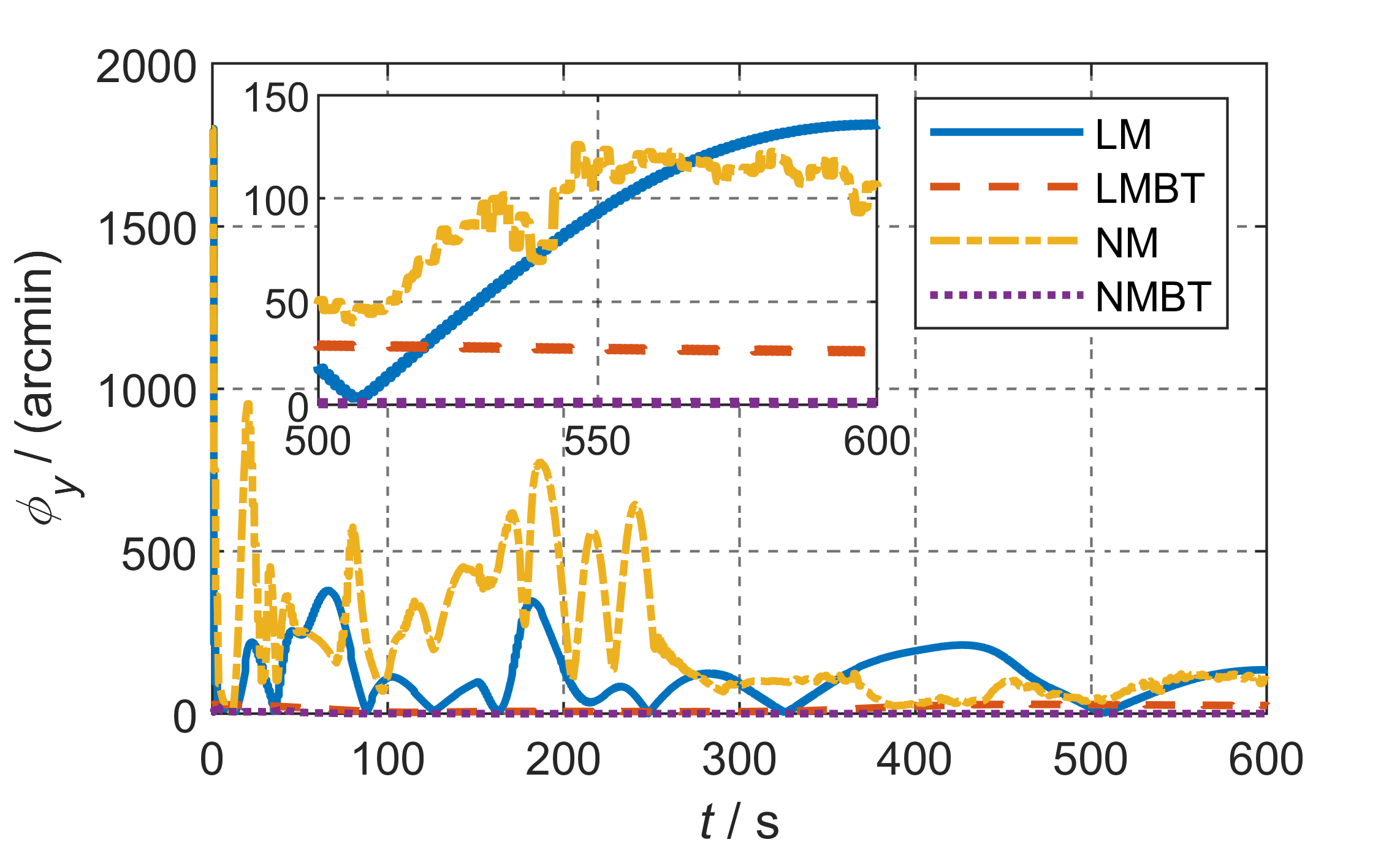}
	\caption{Estimation errors of \textit{y}-axis misalignment angle in simulation}
	\label{fig:sim2_y}
\end{figure}

\begin{figure}[!ht]
	\centering
	\includegraphics[width=\columnwidth]{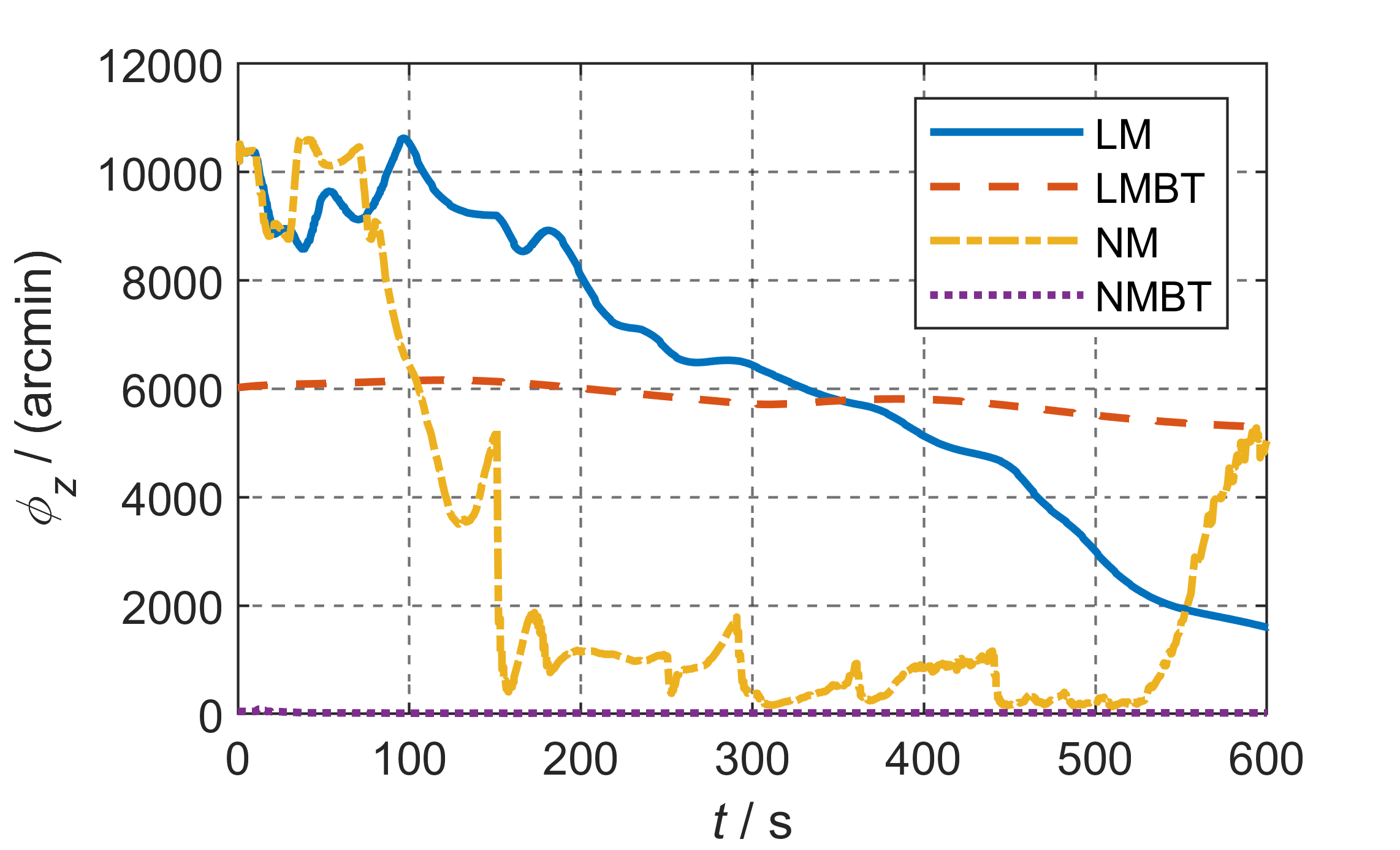}
	\caption{Estimation errors of \textit{z}-axis misalignment angle in simulation}
	\label{fig:sim2_z}
\end{figure}

To test the proposed alignment method under the condition of large misalignment angle, the misalignment angle was set as $ \boldsymbol{\alpha} = [30\degree \quad 30\degree \quad 170\degree] $. The alignment calculation is composed with 3 forward filtering processes and 2 reverse filtering processes. The total alignment time is increased, because the much large misalignment angle need longer filtering time to converge. After 30 times Monte-Carlo simulation finished, the curves of the RMS estimation errors of these algorithms are shown in Figure \ref{fig:sim2_x}, \ref{fig:sim2_y} and \ref{fig:sim2_z}. And the estimation errors in the large misalignment angle condition are listed in Table \ref{tab:simerr2}.


\begin{table}[!ht]
	\centering
	\caption{Estimation error (RMS) of misalignment angle of different alignment algorithms} \label{tab:simerr2}
	\begin{tabular}{lrrr}
		\hline
		Alignment algorithm & $ \phi_x (^\prime) $ & $ \phi_y (^\prime) $ & $ \phi_z (^\prime) $ \\
		\hline
		LMBT   & 	22.5013 &	26.1961 &	5305.1194  \\
		NMBT   &    0.9172  &	1.0547  &	19.8008    \\
		\hline
	\end{tabular}
\end{table}

The large misalignment alignment simulation leads to the following conclusions:
\par (1) Both simple Linear-model alignment and nonlinear-model alignment cannot make three estimation values of misalignment converge to a stable result.
\par (2) Although the backtracking method helps the linear-model alignment to obtain a stable estimation result, the finial misalignment angles are still too large to finish navigation calculation of SINS. Especially, its yaw misalignment is about 90 deg that make the alignment result invalid.
\par (3) The proposed alignment algorithm based on nonlinear model and backtracking algorithm has the highest alignment accuracy. The horizontal misalignment angles are close to $ 1^\prime $, and the heading misalignment angle is less than $ 20^\prime $.

\subsection{Field test on vehicle}

\begin{figure}[!ht]
	\centering
	\includegraphics[width=7cm]{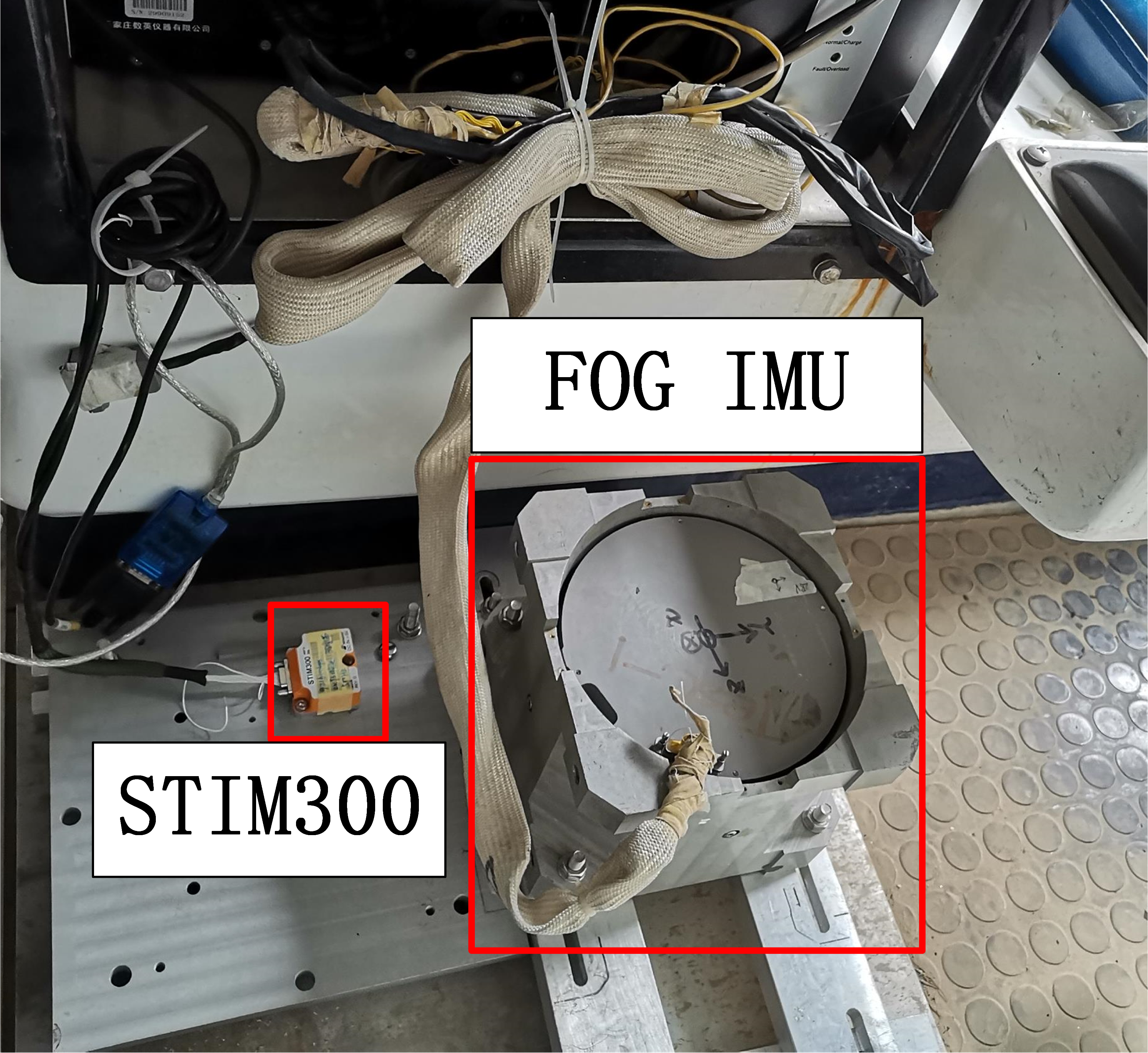}
	\caption{Field test platform}
	\label{fig:test}
\end{figure}

To further test the performance of proposed alignment method based on nonlinear error model and backtracking algorithm, a field test of initial alignment on vehicle was finished with the 4 algorithms listed in Table \ref{tab:sim_4alist}. A STIM-300 MEMS-IMU was used in the test that lasted about 10 minutes (600s). The misalignment angles of coarse alignment were estimated by different algorithm. A high-precision FOG-SINS(Fiber Optic Gyroscope SINS) offered attitude reference for this alignment test by integrating with a high-performance GNSS(1Hz). The parameters of the both IMUs used in test are listed in Table\ref{tab:para}. In addition, the misalignment between MEMS-IMU and FOG-IMU has been measured and compensated through a group of transfer alignment tests finished before this. The field test platform and IMU installation is shown in Fig.\ref{fig:test}.

\begin{table}[!ht]
	\centering
	\caption{Performace parameters of IMUs}\label{tab:para}
	\begin{tabular}{crr}
		\hline
		parameter & STIM-300 & FOG-IMU \\
		\hline
		sample rate						& 125Hz & 200Hz \\
		gyroscope bias				   & $ \le120\degree/\mathrm{h} $ & $ \le0.3\degree/\mathrm{h} $ \\
		gyroscope bias stability     & $ \le0.5\degree/\mathrm{h} $ & $\le0.1 \degree/\mathrm{h} $ \\
		accelerometer bias 					  & $ \le $1mg & $ \le $100ug \\
		accelerometer bias stability  & $\le$100ug & $\le$50ug \\
		\hline
	\end{tabular}
\end{table}

\begin{figure}[!ht]
	\centering
	\includegraphics[width=8cm]{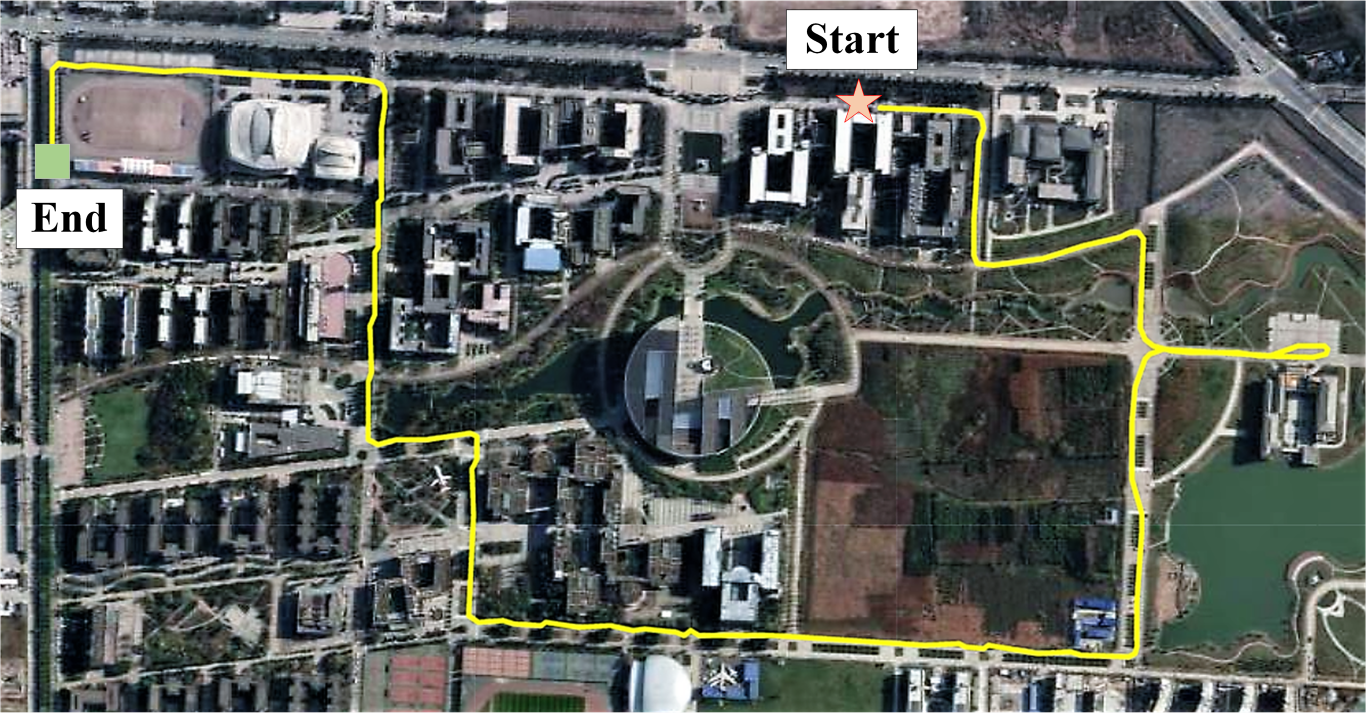}
	\caption{Trajectory of vehicle in field test}
	\label{fig:trj2}
\end{figure}

\begin{figure}[!ht]
	\centering
	\includegraphics[width=8cm]{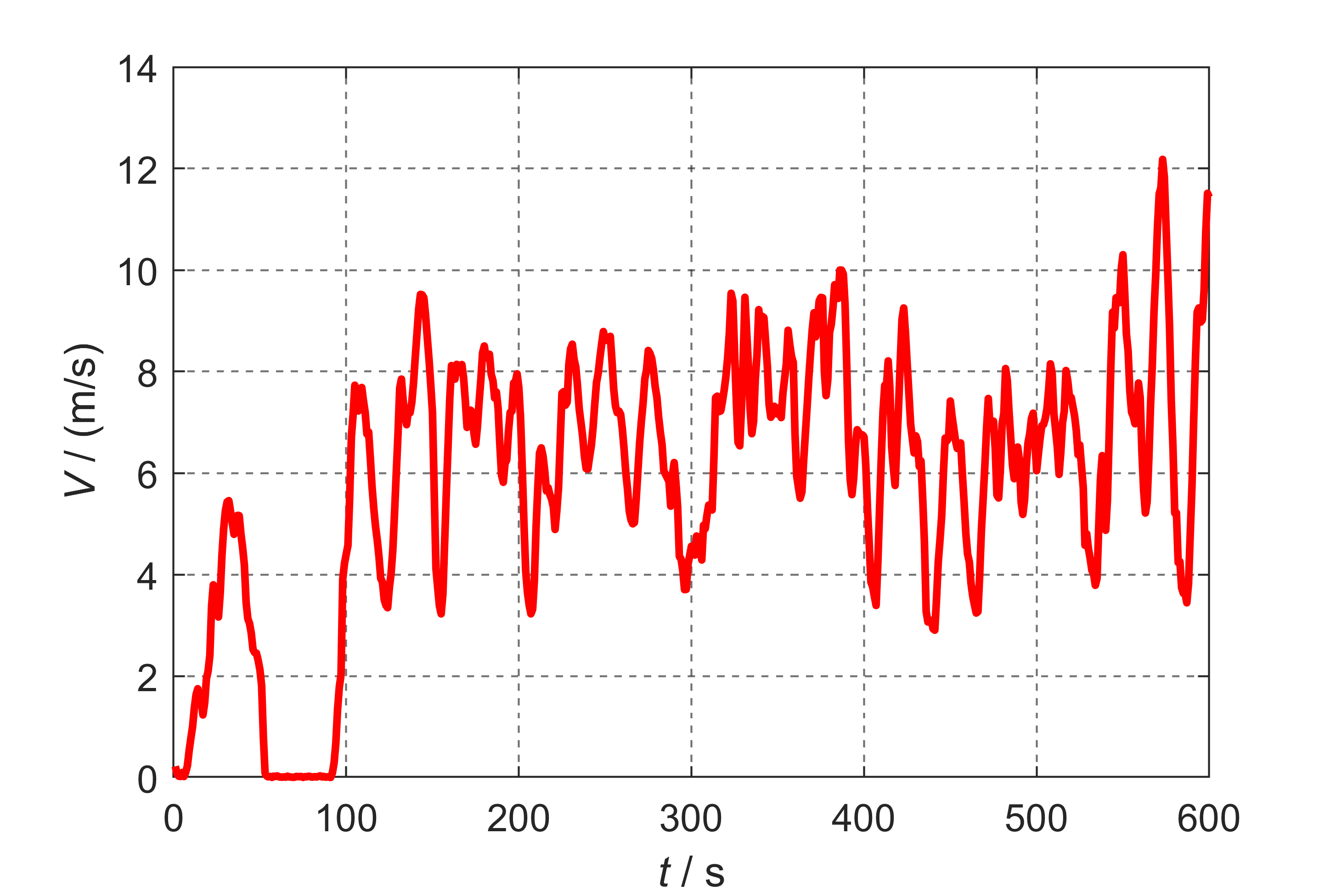}
	\caption{Velocity curve of vehicle}
	\label{fig:tstvn}
\end{figure}

The trajectory of vehicle during test is shown in Figure \ref{fig:trj2}. This test is finished in campus, so the trajectory is composed of many right-angle rotations. The velocity curve of vehicle is shown in Figure \ref{fig:tstvn}. Because the route of test crosses a lot of intersections, vehicle need finish many stop-and-go maneuvers. This two kinds of motion can usefully help Kalman Filter estimating misalignment angles rapidly.

Generally, MEMS-IMU has accurate acceleration output, hence the two horizontal misalignment are small angles. However, the bias error of gyroscope is too large to obtain yaw angle. Hence, the initial misalignment is set as $ \boldsymbol{ \alpha } = [1\degree \quad 1\degree \quad 170\degree]^\mathrm{T} $ to finish this on-field initial alignment algorithms test and evaluate the alignment accuracy of algorithms.

\begin{figure}
	\centering
	\includegraphics[width=\columnwidth]{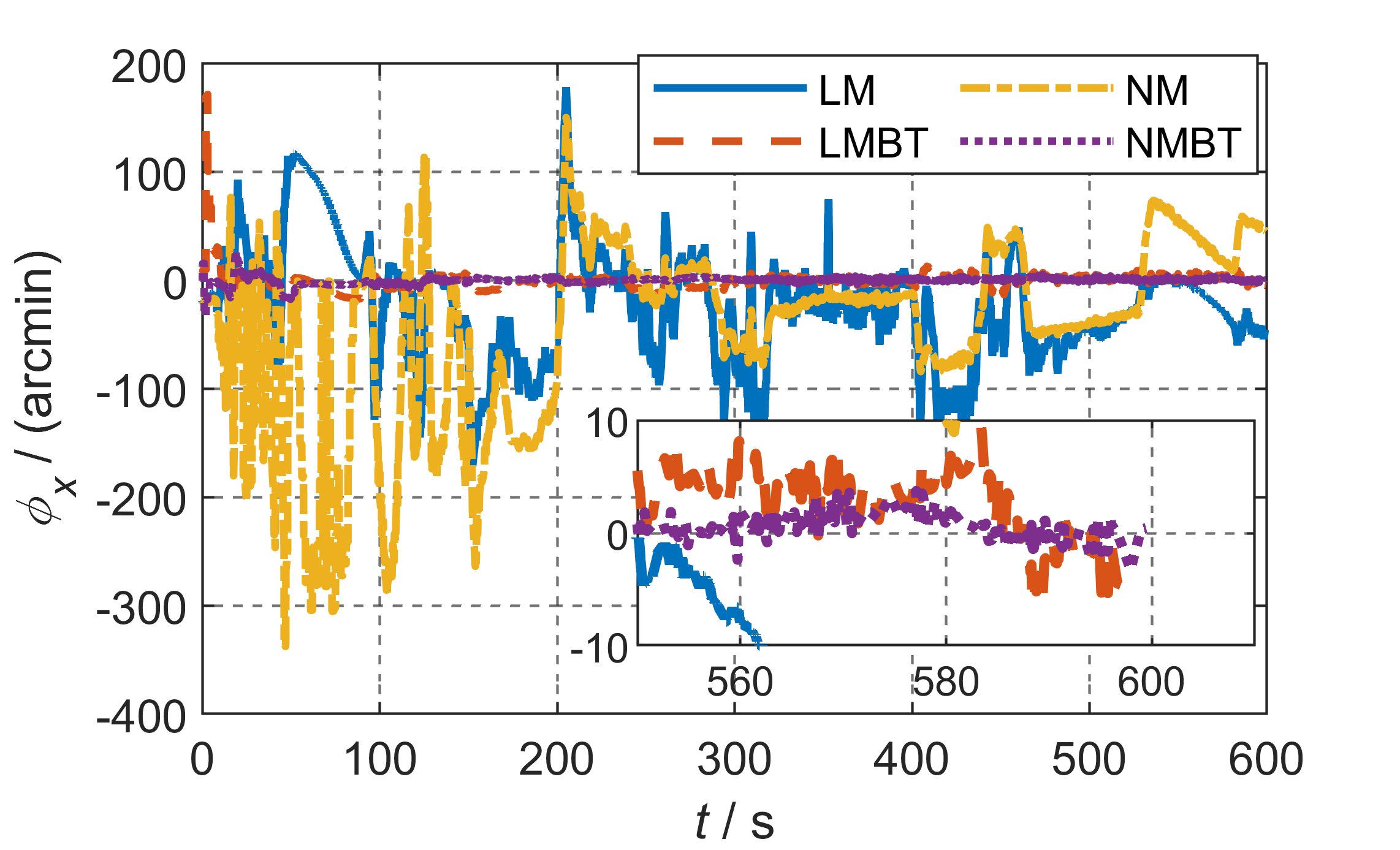}
	\caption{Estimation error of \textit{x}-axis misalignment}
	\label{fig:phix_exp}
\end{figure}

\begin{figure}
	\centering
	\includegraphics[width=\columnwidth]{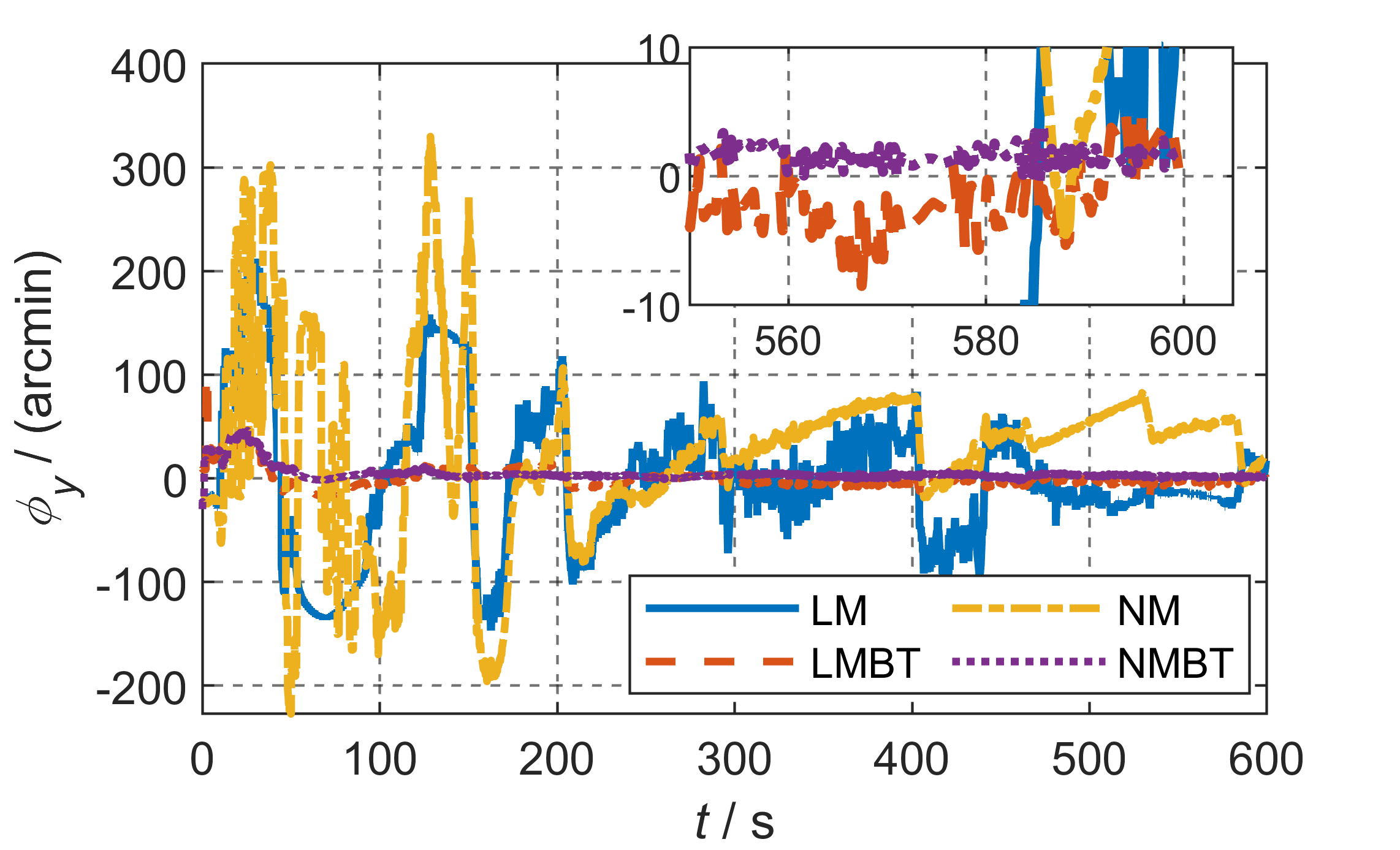}
	\caption{Estimation error of \textit{y}-axis misalignment}
	\label{fig:phiy_exp}
\end{figure}

\begin{figure}
	\centering
	\includegraphics[width=\columnwidth]{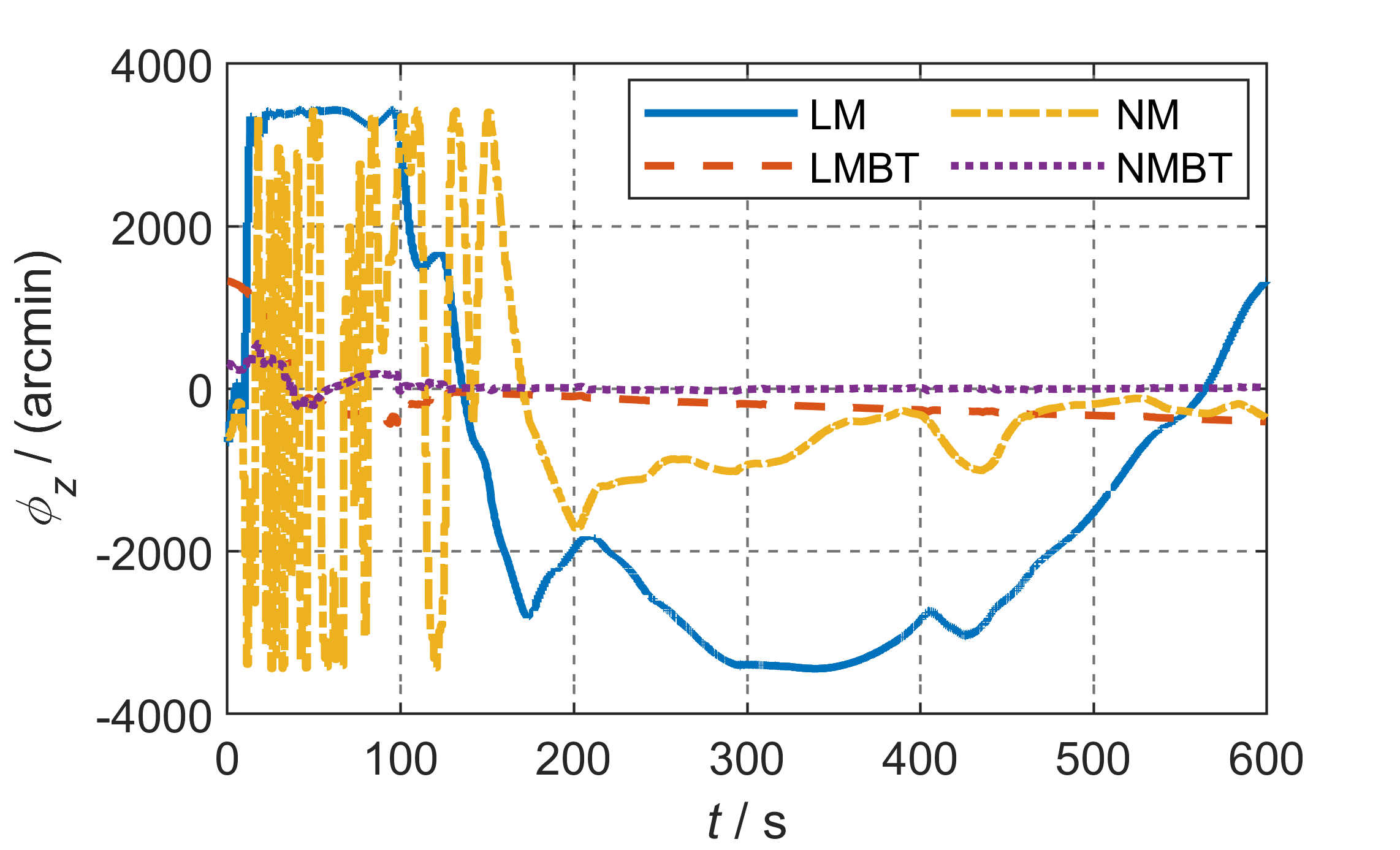}
	\caption{Estimation error of \textit{z}-axis misalignment}
	\label{fig:phiz_exp}
\end{figure}

The misalignment estimation error of alignment algorithms are shown in Figures \ref{fig:phix_exp}, \ref{fig:phiy_exp} and \ref{fig:phiz_exp}. The last 50 seconds misalignment curves of \textit{x} and \textit{y} axes are magnified and located in blank. The RMS results of misalignment angle of last 100s, in which the misalignment estimation values are all convergent in an interval, are calculated and used to evaluate the alignment precision. The alignment errors are listed in Table \ref{tab:experr}.


\begin{table}[!ht]
	\centering
	\caption{Estimation error (RMS) of misalignment angle of different alignment algorithms} \label{tab:experr}
	\begin{tabular}{lrrr}
		\hline
		Alignment algorithm & $ \phi_x (^\prime) $ & $ \phi_y (^\prime) $ & $ \phi_z (^\prime) $ \\
		\hline
	 	LM    & 	34.0522 &	18.9973 &	972.4624  \\
		LMBT  & 	4.1531 &	3.6761 &	357.6490  \\
		NM    & 	43.8879 &	53.2036 &	216.4766  \\
		NMBT  & 	1.3695 &	2.2725 &	10.0261  \\
		\hline
	\end{tabular}
\end{table}

The vehicle navigation alignment test leads to the following conclusions:
\par (1) The proposed alignment algorithm has the best convergence performance and the most accurate alignment result. The two horizontal misalignment angles are $ 1.3695^{\prime} $ and $2.2725^{\prime}$ and misalignment of yaw is only $10.0261^{\prime}$. This result is much better for the low-cost IMUs.
\par (2) The horizontal misalignment angles of the algorithm only using nonlinear error model are worst in these algorithms. The reason is that the distraction to horizontal attitude from the large heading misalignment angle in nonlinear error model is greater than the distraction in linear model. It can be seen by comparing the two attitude error equation.
\par (3) The alignment errors of yaw are sorted as: $ \mathrm{LM}>\mathrm{LMBT}>\mathrm{NM}>\mathrm{NMBT} $. The linear model isn't proper to estimate yaw even the backtracking algorithm is used to aid, when the initial misalignment of yaw is large angle. 
\par (4) The aligning results indicate that the proposed algorithm has great advantage on short-term alignment of SINS in the large misalignment angle condition.

\section{Conclusion}
The low-cost MEMS-IMU that can't obtain the initial attitude independently is wildly applied in integrated navigation system to output precise attitude, velocity and position result. Aligning in moving condition is an important and necessary ability for many vehicles, for an example, the navigation system of tactical missile is started after launching. To solve the problem estimating initial attitude in short term with large misalignment angle, the proposed alignment algorithm, that combines the backtracking method and nonlinear Kalman filter based on the large-misalignment SINS error model, is deduced in detail. Alignment simulations and in-field test indicate that the proposed alignment algorithm has the most accurate aligning result.

%
%
%
\bibliographystyle{IEEEtran}
\bibliography{ref}

\end{document}